\documentclass{article} %
\usepackage{iclr_2023,times}

\usepackage{amsmath,amsfonts,bm}

\def\eqref#1{equation~\ref{#1}}

\def\1{\bm{1}}

\DeclareMathAlphabet{\mathsfit}{\encodingdefault}{\sfdefault}{m}{sl}
\SetMathAlphabet{\mathsfit}{bold}{\encodingdefault}{\sfdefault}{bx}{n}

\usepackage[utf8]{inputenc} %
\usepackage[T1]{fontenc}    %
\usepackage{hyperref}       %
\usepackage{url}            %
\usepackage{booktabs}       %
\usepackage{amsfonts}       %
\usepackage{amsmath}
\usepackage{nicefrac}       %
\usepackage{microtype}      %
\usepackage{xcolor}         %

\usepackage{derivative}
\usepackage{cleveref}
\usepackage{tabu}
\usepackage{caption}

\usepackage[T1]{fontenc}
\usepackage{amsmath}             

\usepackage{amssymb}
\usepackage{mathtools}
\usepackage{amsthm}
\usepackage{pifont}

\newcommand{\cmark}{\ding{51}}%

\usepackage{color, colortbl}
\usepackage{soul}
\usepackage{subcaption}
\usepackage{wrapfig}
\usepackage{wrapfig,lipsum,booktabs}
\usepackage{multirow}
\definecolor{Gray}{gray}{0.9}
\definecolor{darkgray}{rgb}{0.66, 0.66, 0.66}
\definecolor{napiergreen}{rgb}{0.16, 0.5, 0.0}
\definecolor{emerald}{rgb}{0.31, 0.78, 0.47}
\definecolor{electricpurple}{rgb}{0.75, 0.0, 1.0}
\definecolor{yellowtwo}{rgb}{1.0, 0.98, 0.7}
\definecolor{yellowthree}{rgb}{1.0, 0.949, 0.545}

\newcommand{\tabnum}[2]{#1{\tiny (#2)}}
\newcommand{\besttabnum}[2]{\textbf{#1}{\tiny (#2)}}

\DeclareRobustCommand{\hltwo}[1]{{\sethlcolor{yellowtwo}\hl{#1}}}

\title{Substructure-Atom Cross Attention for \\Molecular Representation Learning}

\author{Jiye Kim\thanks{Equal contribution.} \quad Seungbeom Lee$^{*}$ \quad Dongwoo Kim \quad Sungsoo Ahn \quad Jaesik Park\\
Pohang University of Science and Technology~(POSTECH), South Korea\\
\texttt{\small{\{jk3472, slee2020, dongwoo.kim, sungsoo.ahn, jaesik.park\}@postech.ac.kr}
}}

\iclrfinalcopy 
\begin{document}

\maketitle

\crefname{section}{Sec.}{Secs.}
\Crefname{section}{Section}{Sections}
\Crefname{table}{Table}{Tables}
\crefname{table}{Tab.}{Tabs.}

\newcommand{\Eq}[1]  {Eq.\ \ref{eq:#1}}
\newcommand{\Eqs}[1] {Eqs.\ \ref{eq:#1}}
\newcommand{\Fig}[1] {Figure \ref{fig:#1}}
\newcommand{\Figs}[1]{Figures \ref{fig:#1}}
\newcommand{\Tbl}[1]  {Table \ref{tbl:#1}}
\newcommand{\Tbls}[1] {Tables \ref{tbl:#1}}
\newcommand{\Sec}[1] {Section \ref{sec:#1}}
\newcommand{\Secs}[1] {Sections \ref{sec:#1}}
\newcommand{\App}[1] {Appendix \ref{app:#1}}
\newcommand{\Alg}[1] {Algorithm \ref{alg:#1}}
\newcommand{\Etal}   {et al.}

\newcommand{\OursAcronym}{WS${}^3$-ConvNet}

\newcommand{\jh}[1]{{\textcolor{orange}{JH: #1}}}
\newcommand{\kh}[1]{{\textcolor{blue}{KH: #1}}}
\newcommand{\todo}[1]{{\textcolor{red}{#1}}}

\newcommand{\minus}[1]{{\textcolor{purple}{$\downarrow$ #1}}}
\newcommand{\plus}[1]{{\textcolor{teal}{$\uparrow$ #1}}}
\newcommand{\nminus}[1]{{\textcolor{teal}{$\downarrow$ #1}}}
\newcommand{\nplus}[1]{{\textcolor{purple}{$\uparrow$ #1}}}

\newcommand{\aminus}[1]{{\textcolor{purple}{#1}}}
\newcommand{\aplus}[1]{{\textcolor{teal}{#1}}}
\newcommand{\anminus}[1]{{\textcolor{teal}{#1}}}
\newcommand{\anplus}[1]{{\textcolor{purple}{#1}}}
\newcommand{\mathMesh}{\mathcal{M}}
\newcommand{\mathFace}{f}
\setlength{\textfloatsep}{0.8cm}
\setlength{\floatsep}{0.8cm}

\begin{abstract}
\label{abstract}
Designing a neural network architecture for molecular representation is crucial for AI-driven drug discovery and molecule design. In this work, we propose a new framework for molecular representation learning. Our contribution is threefold: (a) demonstrating the usefulness of incorporating substructures to node-wise features from molecules, (b) designing two branch networks consisting of a transformer and a graph neural network so that the networks fused with asymmetric attention, and (c) not requiring heuristic features and computationally-expensive information from molecules. Using 1.8 million molecules collected from ChEMBL and PubChem database, we pretrain our network to learn a general representation of molecules with minimal supervision. The experimental results show that our pretrained network achieves competitive performance on 11 downstream tasks for molecular property prediction. 

\end{abstract}

\section{Introduction}
\label{introduction}

Predicting properties of molecules is one of the fundamental concerns in various fields. For instance, researchers apply deep neural networks (DNNs) to replace expensive real-world experiments to measure the molecular properties of a drug candidate, e.g., the capability of permeating the blood-brain barrier, solubility, and affinity. Such an attempt significantly reduces wet-lab experimentation that often takes more than ten years and costs \$1 million \citep{hughes2011principles, mohs2017drug}.

Among the DNN architectures, graph neural networks (GNNs) and Transformers are widely adopted to recognize graph structure of molecules. GNN architectures for molecular representation learning include message-passing neural network (MPNN) and directed MPNN~\citep{gilmer2017neural, yang2019analyzing}, where they investigate how to obtain effective node,  edge, and graph representation. GNNs are powerful in capturing local information of a node, but may lack the ability to encode information from far-away nodes due to over-smoothing and over-squashing issues~\citep{li2018deeper,alon2020bottleneck}.
On the other hand, Transformer-based architectures such as MAT~\citep{maziarka2020molecule} and Graphormer~\citep{ying2021transformers} augment the self-attention layer of a vanilla Transformer using high-order graph connectivity information. Transformers can encode global information as they consider attention between every pair of nodes from the first layer.
To guide a structural bias in the attention mechanism, previous work relies on heuristic features such as the shortest path between two nodes since the naive Transformers cannot recognize the graph structure.

From the understanding of chemical structure, it is known that meaningful substructures can be found across different molecules, also known as motif or fragments \citep{murray2009rise}.
For example, carbon rings and NO2 groups are typical substructures contributed to mutagenicity~\citep{debnath1991structure} showing that proper usage of substructures can help a property prediction.
Molecular substructures are often represented as molecular fingerprints or molecular fragmentation. Molecular fingerprints such as MACCS (Molecular ACCess System) keys~\citep{durant2002reoptimization} and Extended-Connectivity Fingerprints (ECFPs)~\cite{rogers2010extended} represent a molecule into a fixed binary vector where each bit indicates the presence of a certain motif in the molecule. With a predefined fragmentation dictionary, such as BRICS~\citep{degen2008art} or tree decomposition~\citep{jin2018junction}, a molecule can be decomposed into distinct partitions. Interestingly, machine learning algorithms that utilize molecular substructures still show competitive performance on some datasets to deep learning models~\citep{hu2020ogb, maziarka2020molecule}.

\begin{figure}[t!]
    \centering
    \begin{subfigure}[b]{0.48\linewidth}
        \includegraphics[width=\linewidth]{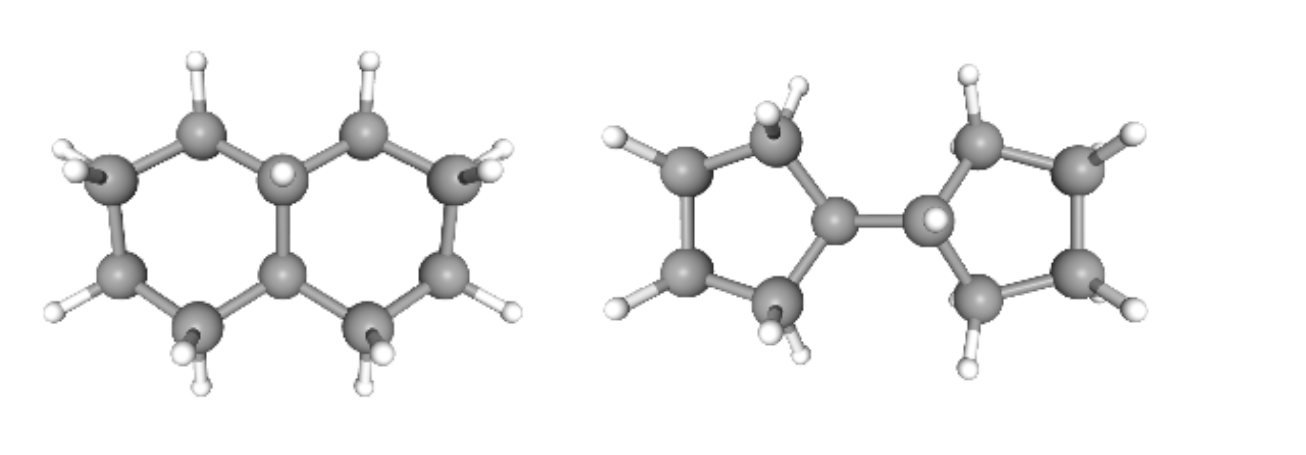}
        \caption{Indistinguishable molecules by traditional GNNs }
        \label{fig:graph_shortcoming}
    \end{subfigure}
    \hfill
    \begin{subfigure}[b]{0.48\linewidth}
        \includegraphics[width=\linewidth]{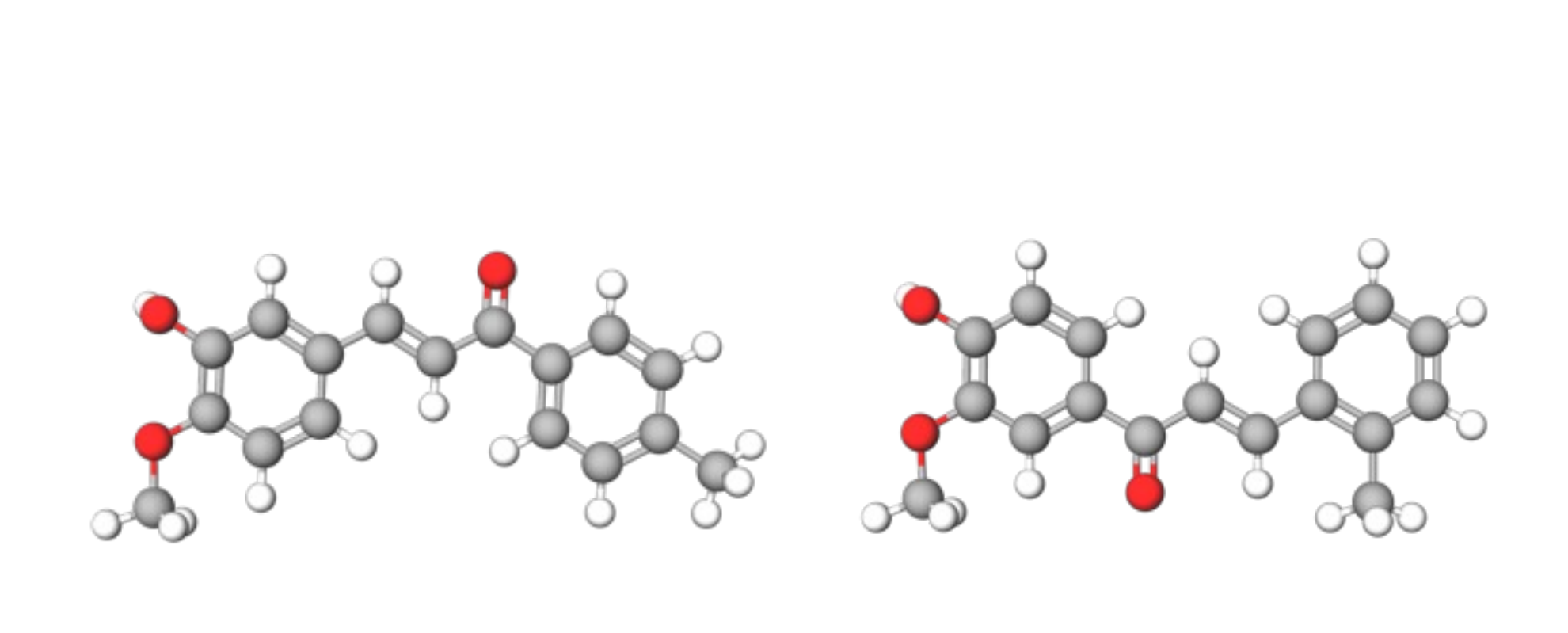}
        \caption{Indistinguishable molecules by MACCS keys}
        \label{fig:tf_shortcoming}
    \end{subfigure}
    \caption{Traditional GNNs, such as GCN and GAT, cannot distinguish the two molecular graphs in (a). However, it can be easily distinguished through simple 6-ring and 5-ring substructure information. On the other hand, the two molecules in (b) have similar substructures, so atom features from neighborhood are necessary to discriminate the two molecules.}
    \label{fig:intro}
\vspace{-.2in}
\end{figure}

We propose a fusion architecture between a GNN and Transformer to  incorporate molecular graph information and molecular substructures.  Molecular substructures and graph are encoded through Transformer and GNN, respectively. The Transformer is designed to recognize the molecular substructures, where the substructure information is mixed through self-attention to obtain better representations. With the Transformer only architecture, however, local information of molecules, such as atoms, bonds, and connectivity, can be lost from the structures. For example, the two molecules shown in \autoref{fig:tf_shortcoming} share the same representation with MACCS keys while having different structures.
To overcome, we use a separate GNN branch for preserving local information. In our model, we inject the GNN feature into the intermediate Transformer layers through the fusing network. In this way, substructures and local node information are interactively fused, producing a final representation for molecular graphs. 

We name our network as \emph{\textbf{Substructure-Atom Cross Attention (SACA)}} as it uses substructure as well as atom information in molecules and fuses them through cross-attention.
The architectural choices allows us to avoid heuristic features such as the shortest-path or 3-dimensional distance in attention layers, which are required in many existing Transformer architectures for molecular representation learning~\citep{maziarka2020molecule, ying2021transformers}.
Furthermore, our model reduces the complexity for attention calculation over the node-level Transformer models from $O(N^{2})$ to $O(N)$, where $N$ is the number of atoms.
To demonstrate the empirical effectiveness of the proposed network and see the ability to capture the general representation of molecules, we evaluate our model on 11 downstream tasks from MoleculeNet~\citep{wu2018moleculenet}. Our approach achieves the competitive performance on 11 downstream tasks. 

In what follows, we summarize the key contributions and benefits.
\begin{itemize}
    \item{We propose a novel network that combines the information from substructures and node features in a molecule. Our model combines the advantages of both Transformer and GNN architectures to represent the information given to each architecture.}
    \item{We show the effectiveness of our model for molecular representation learning. Our model achieves competitive performance upon strong baseline models on 11 molecular property tasks.}
    \item{Our model does not require computationally-expensive heuristic information of molecular graphs.}
\end{itemize}

\section{Related Work}
\label{related work}

\subsection{Architectures for Molecular representation learning}

\paragraph{Graph neural networks (GNNs)}

Most common architecture for molecular representation learning is the GNNs since molecules can be naturally represented as a graph structure; a node as an atom, an edge as a connection. %
Researchers have actively investigated variations of GNN architectures \citep{gilmer2017neural, yang2019analyzing, xiong2019pushing, song2020communicative},  for molecules. For example, MPNN~\citep{gilmer2017neural} generalizes the message passing frameworks and explores some variants that predict molecular properties. Directed MPNN (DMPNN) \citep{yang2019analyzing} proposed to replace the node-based message by edge-based messages to avoid unnecessary message loops. Next, communicative MPNN (CMPNN) \citep{song2020communicative} improved DMPNN by additionally considering the node-edge interaction during the message passing phase. AttentiveFP~\citep{xiong2019pushing} extends the graph attention mechanism to allow for nonlocal effects at the intramolecular level.

Despite the advance of GNN architectures for molecular representation learning, there are known problems in GNN, such as over-smoothing and over-squashing problems~\citep{li2018deeper,alon2020bottleneck}, which means the node representations become too similar, and the information from far nodes does not propagate well as the number of neighbors increases exponentially. Furthermore, \citet{xu2018powerful} analyzes expressive powers of standard GNNs following neighborhood aggregation scheme and shows that they are bounded to Weisfeiler-Lehman test (WL-test)~\citep{weisfeiler1968reduction}. Therefore, GNNs with standard message passing cannot learn to discern a simple substructure such as  cycles. For example, the two molecules in \ref{fig:graph_shortcoming} cannot be distinguished by WL-test, hence standard GNNs cannot distinguish these two molecules~\citep{bodnar2021weisfeiler}. A solution to this limited representation power of GNNs is to directly incorporate important substructures in the representation learning framework.

\vspace{-2mm}

\paragraph{Transformers}
With recent advance of Transformer architectures and their promising performance in various domains, including NLP and computer vision~\citep{devlin-etal-2019-bert,dosovitskiy2020image}, Transformer-based architectures for molecular representation learning~\citep{maziarka2020molecule,ying2021transformers,rong2020grover, maziarka2021relative} have been developed. Transformer architecture calculates pair-wise attention between every node from the first layer. Therefore, it can effectively capture the global information of a graph. However, a vanilla Transformer architecture~\citep{vaswani2017attention} is not directly applicable for molecular graph representation because it cannot incorporate structural information such as the edge and connectivity in graphs. To bridge the gap, advanced Transformer architectures alter the self-attention layer~\citep{maziarka2020molecule,ying2021transformers, maziarka2021relative} or incorporate message passing networks into Transformer architectures for input feature~\citep{rong2020grover}. Specifically, MAT~\citep{maziarka2020molecule} uses adjacency and distance matrix of atoms to augment the self-attention layer. GROVER~\citep{rong2020grover} runs Dynamic Message Passing Network (dyMPN) over the input node and edge features to extract queries, keys and values for self-attention layers. CoMPT~\citep{chen2021learning} uses the shortest path information between two nodes, and Graphormer~\citep{ying2021transformers} encodes node's degree, edges and the shortest path with edges between two nodes for molecular data. Although these graph-specific features such as the shortest path or 3D distance are found to be useful in graph representation, these features can be heuristic and impose excessive computational overhead, which limits the model's applicability to large molecules such as proteins. 
Our proposed model does not require any computationally expensive heuristic features such as 3D information or shortest path for preprocessing or computation in attention layer.

\subsection{Molecular Substructure}

Substructure information extracted from molecules has been widely used in molecular generation, property prediction, and virtual screening~\citep{willett1998chemical, brown1996use, eckert2007molecular, jin2018junction}. ECFPs~\citep{rogers2010extended} encode existing substructures within a circular distance from each atom in a molecule. PMTNN~\citep{ramsundar2015massively} is a multi-task network that takes ECFPs as an input to predict molecular properties on various datasets. MACCS keys~\citep{durant2002reoptimization} extract molecules substructure depending on the presence of pre-defined functional groups. One example of the usage of MACCS keys is to encode known ligands of each protein, which leads to an improvement of prediction performance in protein-ligand interaction~\citep{li2019predicting}. Molecular fragmentation method such as BRICS fragmentation~\citep{degen2008art, zhang2021motif} or tree decomposition~\citep{jin2018junction} is to decompose molecules in non-overlapping partitions with pre-defined rules. BRICS fragmentation divides molecules following chemical reaction based rules. Tree decomposition proposed in~\cite{jin2018junction} also extracts junction tree by contracting certain edges. One node in the junction tree represents a substructure of original molecules. Our model receives molecular substructures as input to supplement GNN's expressivity and it can flexibly encode any substructure vocabulary.

Our model combines Transformer and GNN, with molecular substructures and molecular graph as inputs to each network. There are existing studies that have also considered combining GNN and Transformer architecture~\citep{zhu2021dual, zhu2021posegtac, yang2021graphformers} to learn graph representations. Specifically, DMP~\citep{zhu2021dual} utilizes GNN and Transformer to encode graphs and SMILES representation of molecules and train both branches using a consistency loss to match the two outputs of input molecules. PoseGTAC~\citep{zhu2021posegtac} and GraphFormers~\citep{yang2021graphformers} combine GNN and Transformer layer alternatively to enlarge the receptive field or to mix the output of Transformer. Our model is the first to encode substructures through Transformer and inject atom features through a separate GNN. In this way, we preserve both substructures and local atom features of molecules.

\section{Model}
\label{method}
\begin{wrapfigure}{br}{0.55\linewidth}
    \begin{minipage}{1.0\linewidth}
\vspace{-2em}
    \includegraphics[width=\linewidth]{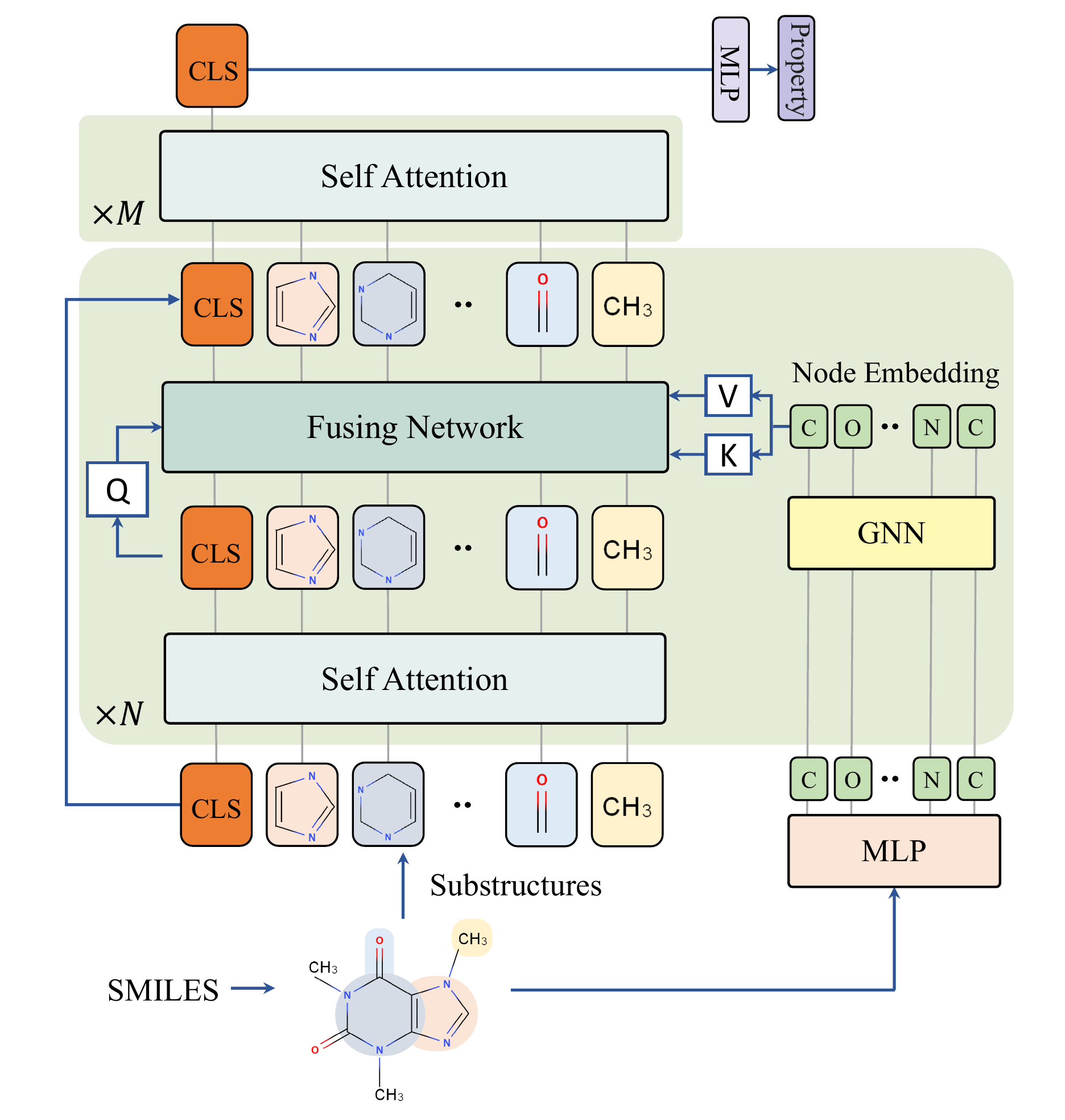}
    \caption{Overall architecture. Our model consists of Transformer and GNN branches. Transformer encodes molecular substructures, while GNN encodes atomic information. The self-attention and fusing network are alternatively stacked $N$ times, followed by $M$ more self-attention layer to refine the substructure feature. The final CLS token is used to predict the molecular properties.}
\label{fig:overall}
\vspace{-2em}
    \end{minipage}
\end{wrapfigure}

In this section, we explain the architecture of our model.

\paragraph{Overall architecture}
\label{overall_architecture}
\autoref{fig:overall} shows the overall architecture of our model. Our network consists of two branches: (i) a Transformer branch that uses the molecular substructures as input and (ii) a GNN branch that uses a molecular graph as input. The two branches have different roles. First, the Transformer branch is intended to capture global information of molecules. It receives the molecular substructures that have important role in molecular properties, but cannot be easily captured by GNNs, and learn the overall representation of molecules. On the other hand, the GNN branch is intended to capture local node information of molecules. The two different levels of information are mixed through the fusing network in the Transformer branch.

\paragraph{Transformer branch}
Our Transformer branch is to incorporate both molecular substructure information and local node features. The input token for Transformer is the substructure embeddings of molecules. Predefined substructures are first detected and then projected into separate embedding vectors. For example, in \autoref{fig:overall}, substructures such as N-Heterocycle, carbon-oxygen bond and methyl group are detected and embedded into learnable embedding vectors. To identify substructure from the input molecules, we use MACCS keys~\citep{durant2002reoptimization}, which indicates the presence of motifs in a molecule. Note that our architecture is not limited to certain molecular substructures, but it can flexibly receive any substructure vocabulary. The embeddings of substructures are mixed together and refined as they are passed through the self-attention module.

\begin{wrapfigure}{l}{0.5\linewidth}

\centering
  \begin{minipage}{1.0\linewidth}
    \includegraphics[width=1.0\textwidth]{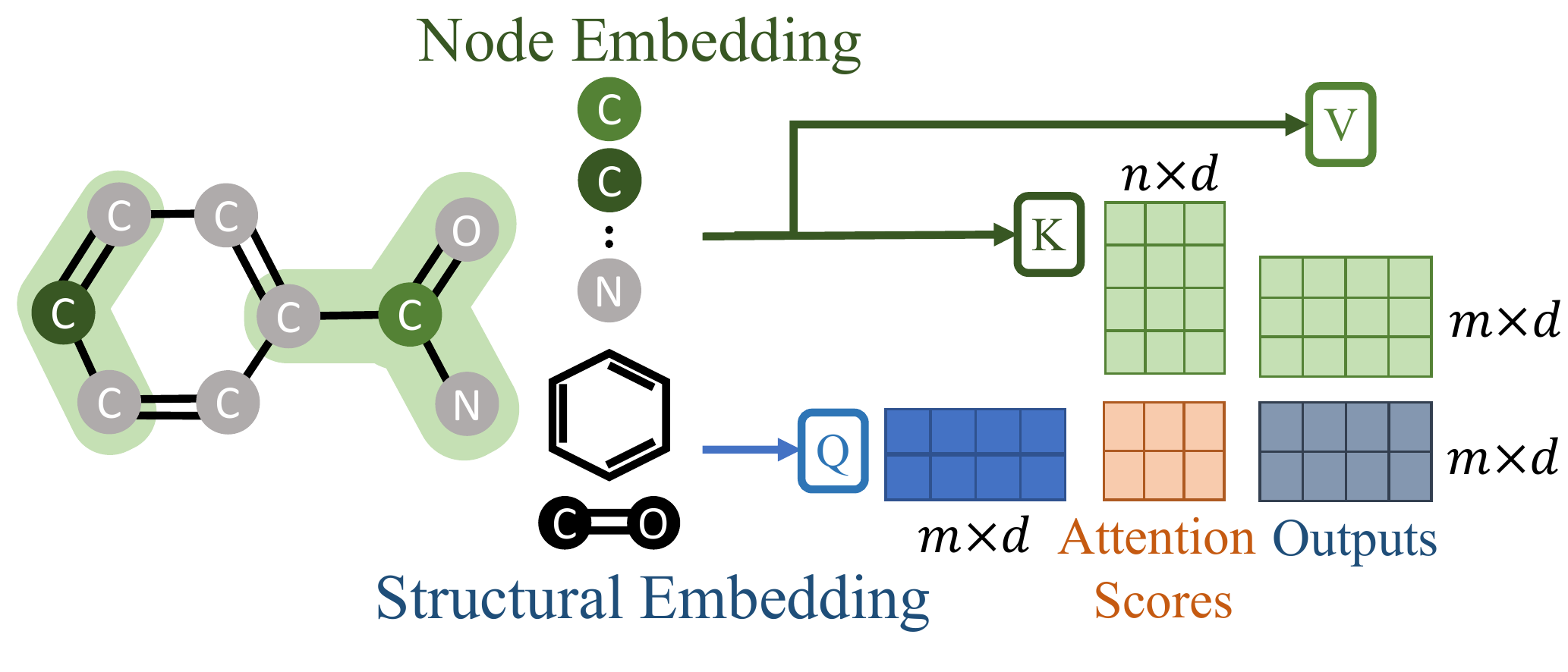}
    \vspace{-5mm}
  \caption{Illustration of structural and node embedding for cross attention computation. Cross attention is computed with the illustrated query, key and value.}
  \label{fig:cross-attention}
  \end{minipage}
 \vspace{-5pt}

\end{wrapfigure}

The substructure embeddings after self-attention layer are fused with node embeddings from a separate GNN branch. The fusing network computes cross-attention between substructures and nodes where substructures are used as query and nodes are used as key and value. The detailed computation of the fusing network is shown in \autoref{fig:cross-attention}. In the fusing network, the cross-attention between each pair of substructure embedding and node embedding is computed.

To be specific, for a given molecule having $n$ atoms and $m$ extracted substructures, we have substructure embeddings $E_s \in \mathbb{R}^{m \times d}$ and node embeddings $E_n \in \mathbb{R}^{n \times d}$ where $d$ is the embedding dimension. 
Then, the cross-attention is computed as follows:
\begin{equation}
\operatorname{Attention}(Q, K, V) = \operatorname{Softmax}\left(\frac{(E_sW_Q)(E_nW_K)^T}{\sqrt{d_k}}\right)(E_nW_V),
\label{eqn:cc}
\end{equation}
where $W_Q, W_K, W_V \in \mathbb{R}^{d \times d_k}$ are learnable parameters.
The cross-attention module outputs $E_s' \in \mathbb{R}^{m \times d_k}$. Through this fusing network, the substructure embeddings aggregate the local information from node embeddings. Structurally important nodes are aggregated with more weights. Instead of designing heuristic weights on the nodes, our model can learn to select structurally important nodes related to graph-level property by cross-attention. Additionally, as the attention is computed between substructures and nodes, the space and time complexity of the self and cross attention map of our model is linear to the number of atoms, i.e., $O(N)$, whereas other transformer architectures for molecular graphs have quadratic complexity.

The self-attention between substructures and fusing network with node embeddings are repeated iteratively $N$ times. Before making the final prediction, we add $M$ self-attention layers at the end of the network for making refinement on the substructures. We add a residual connection from the input tokens to every output of the fusing network. This ensures the input structural information last throughout the entire network. Furthermore, we use a CLS token similar to special classification token in BERT~\citep{devlin-etal-2019-bert} to aggregate the global representation of molecules. The CLS token is shown in \autoref{fig:overall}. It is a learnable latent vector and passed to the Transformer branch attached to the substructure tokens as input. A CLS token has been used in many Transformer-based architectures~\citep{maziarka2020molecule, ying2021transformers} for molecular representation learning. We attach an MLP head to the final CLS token to perform graph-level prediction tasks.

\paragraph{GNN branch}
The GNN branch is used to extract local node features from molecular graphs. For GNN architecture, we use GIN~\citep{xu2018powerful} with jumping knowledge~\citep{xu2018representation}. The computed node features are injected into the Transformer through the fusing network with the same hierarchy. For example, 0-hop node representations are injected into the first cross-attention module and 1-hop representations for the second cross-attention module. This allows the model to encode local node features progressively from the shallow to deeper layers.

The computation of node representation through GNNs and injection to the Transformer allow us to take advantage of both GNNs and Transformer architectures. Substructures that are hard to be captured by GNNs, but essential to molecular properties, are first detected and encoded through Transformer. Meanwhile, local node information that can be lost in using substructures alone is effectively captured by GNNs and fused with substructures. Additionally, our architecture does not require computationally expensive high-order graph-level information. As Transformer cannot naturally incorporate a graph's edge connectivity information, existing work~\citep{rong2020grover, ying2021transformers,chen2021learning} mainly focuses on how to add structural bias such as the shortest path or 3D distance between two nodes into the Transformer self-attention computation. However, these structural biases require computationally expensive preprocessing of the molecular datasets. Our network that utilizes GNN as a separate branch can avoid these limitations.

\section{Experiment}
\label{experiment}

\subsection{Experimental Setting}
\label{exp_setting}
\paragraph{Pretraining}
We pretrain our network to obtain molecular representation transferable to various molecular datasets and tasks. For pretraining, we extracted 200 real-valued descriptors of physicochemical properties from the pretraining datasets using RDKit~\citep{Landrum2016RDKit2016_09_4} and train our network to predict these properties. As the 200 molecular descriptors include a diverse set of molecular properties, the model can learn a representation of molecules that can be used for various downstream tasks.

\paragraph{Dataset}
We collected 1,858,081 number of unlabeled molecules from ChEMBL and PubChem databases~\citep{kim2016pubchem, gaulton2012chembl}. ChEMBL and PubChem are large-scale databases that include a variety of chemical and physical properties, and biological activities of molecules. To obtain molecular substructures, we utilize MACCS key~\citep{durant2002reoptimization}, a 166 dimensional vector that indicates a presence of certain substructure in molecules, and extract this for every molecule using RDKit~\citep{Landrum2016RDKit2016_09_4}. We use OGB package~\citep{hu2020ogb} to convert SMILES~\citep{weininger1988smiles}, a text-representation for molecules, to molecular graphs. 

\paragraph{Implementation details}
We set $M= 4$ and $N=3$, where $M$ and $N$ are defined in~\autoref{overall_architecture}. We set 768-dimensional hidden units and 16 attention heads. When pretraining, we used the AdamW optimizer~\citep{loshchilov2018decoupled} with a learning rate of 1e-4. We divide the pretraining dataset into a 9:1 ratio and use them for training and validation sets. The model is trained for 10 epochs, and the model with the best validation loss is used for downstream tasks. Further details for pretraining setting is presented in \autoref{appendix:experimental setting}.

\subsection{Downstream tasks}

\paragraph{Tasks}
We evaluate the performance on six classification tasks (BBBP, SIDER, ClinTox, BACE, Tox21, and ToxCast) and five regression tasks (FreeSolv, ESOL, Lipo, QM7 and QM8) from MoleculeNet~\citep{wu2018moleculenet}.
The dataset statistics are summarized in \autoref{tab:datasets_info}. Each task is related to molecular property from low-level, for example, water solubility in ESOL to high-level, possibility of blood-brain barrier penetraion in BBBP. Further details about the downstream tasks are available in \autoref{appendix:downstream_dataset}.

\begin{table}[!t]
\centering
\caption{Statistics of eleven datasets from MoleculeNet~\citep{wu2018moleculenet} used for downstream tasks.}
\label{tab:datasets_info}
\begin{subtable}[h]{0.45\linewidth}
    \caption{Classification tasks}
    \resizebox{\linewidth}{!}{
    
    \begin{tabu}{lccc}
    \toprule
    Dataset & Size & \# Tasks & Metric
    \\ 
    \midrule
    BBBP& \phantom{00}2,039 & \phantom{00}1   & ROC-AUC
    \\
    SIDER & \phantom{00}1,427   & \phantom{0}27      & ROC-AUC
    \\
    ClinTox & \phantom{00}1,478   & \phantom{00}2         & ROC-AUC
    \\
    BACE & \phantom{00}1,513  & \phantom{00}1     & ROC-AUC
    \\
    Tox21 & \phantom{00}7,831 & \phantom{0}12  & ROC-AUC
    \\
    ToxCast &  \phantom{00}8,575      & 617    & ROC-AUC
    \\
    \bottomrule    
    
    \end{tabu}
}
\end{subtable}
\quad
\begin{subtable}[h]{0.45\linewidth}
    \caption{Regression tasks}
    \resizebox{\linewidth}{!}{
    
    \begin{tabu}{lccc}
    \toprule
    Dataset & Size & \# Tasks  & Metric
    \\ 
    \midrule
    FreeSolv & \phantom{00,}642 & \phantom{0}1 & RMSE
    \\
    ESOL & \phantom{0}1,128   & \phantom{0}1  & RMSE
    \\
    Lipophilicity & \phantom{0}4,200   & \phantom{0}1  & RMSE
    \\
    QM7 & \phantom{0}6,830  & \phantom{0}1  & MAE
    \\
    QM8 & 21,786 & 12  & MAE
    \\
    \bottomrule
    \vspace{2px}
    \end{tabu}
}
\end{subtable}
\vspace{-1em}
\end{table}

\paragraph{Experimental setting}
To evaluate each model, we use 3 different scaffold splits~\citep{wu2018moleculenet} following~\cite{rong2020grover}. Scaffold split divides structurally different molecules into different subsets and provides more challenging and realistic test environment. 
From the pretrained model, we replace the last MLP layer of the network with the task-specific MLP heads. For each downstream dataset, we train our model for 100 epochs and report the test score corresponding to the best validation epoch. We tune hyperparameters with Bayesian optimization search with a budget of 100 for learning rate, dropout, weight decay and the number of last prediction heads. The hyperparameter search range is provided in \autoref{appendix:experimental setting}.

We compare the performance of our model on the downstream tasks with several GNN and Transformer based state-of-the-arts approaches for molecule representation learning. TF-Robust~\citep{ramsundar2015massively} is a DNN-based model that takes molecular fingerprints. GNN-based models include GraphConv~\citep{kipf2016semi}, Weave~\citep{kearnes2016molecular} and SchNet~\citep{schutt2017schnet} which are 3 graph convolutional networks, MPNN~\citep{gilmer2017neural}, DMPNN~\citep{yang2019analyzing}, MGCN~\citep{lu2019molecular} and CMPNN \citep{song2020communicative} which are GNN models considering the edge features during message passing. AttentiveFP~\citep{xiong2019pushing} is an extension of graph attention network for molecule representation. Transformer-based models include GROVER \citep{rong2020grover}, MAT~\citep{maziarka2020molecule}, Graphormer~\citep{ying2021transformers} and CoMPT~\citep{chen2021learning}. Among the baselines, N-GRAM~\citep{liu2019n}, \citet{hu2019strategies}, GraphLoG~\citep{xu2021self}, MAT, GROVER, Graphormer, GEM~\citep{fang2022geometry} and MPG~\citep{li2021effective} are models that use pretraining strategies. We report GROVER base model for a fair comparison in terms of the number of parameters. To reproduce the results for models using pretraining strategies, we use the pretrained model made available by the authors. We reproduce the results of MPG due to the different splits used in~\cite{li2021effective}\footnote{Note that there is a mismatch between the splits used in the original paper of MPG and the one used in the author's repository. For both cases, our model performs better than MPG under the same splits. The additional experiments are available in \autoref{appendix:mpg}.}. 
We also evaluate our model with the same data split used for MPG and the results can be found in the appendix. %

\begin{table*}[t]

\begin{center}

\caption{Comparison on small-scale datasets. We report the average and standard deviation (in brackets) over three splits. We mark the best and the second-best performances in \hl{\textbf{bold yellow}} and \hltwo{light yellow}, respectively. The baseline results except for MAT, Graphormer, CMPNN, CoMPT, GraphLoG, GEM and MPG are taken from~\citet{rong2020grover}.}
\label{tab:results}
\caption*{Classification Tasks}
\vspace{-1.2em}

\resizebox{\textwidth}{!}{
\begin{tabular}{lccccccccc}
\midrule[1.0pt]
 Method & Pre. & BBBP~$\uparrow$ & SIDER~$\uparrow$ & ClinTox~$\uparrow$ & BACE~$\uparrow$ & Tox21~$\uparrow$ & ToxCast~$\uparrow$ & Rank
\\ 
\midrule[0.5pt]
 TF\_Robust~{\small\citep{ramsundar2015massively}}& - &\tabnum{\phantom{0}.860}{.087} & \tabnum{\phantom{0}.607}{.033} & \tabnum{\phantom{0}.765}{.085}& \tabnum{\phantom{0}.824}{.022} & \tabnum{\phantom{0}.698}{.012} & \tabnum{\phantom{0}.585}{.031} & 15.0
\\
 Weave~{\small\citep{kearnes2016molecular}}& - &\tabnum{\phantom{0}.837}{.065} & \tabnum{\phantom{0}.543}{.034} & \tabnum{\phantom{0}.823}{.023}& \tabnum{\phantom{0}.791}{.008} & \tabnum{\phantom{0}.741}{.044} & \tabnum{\phantom{0}.678}{.024} & 16.0
\\
 GraphConv~{\small\citep{kipf2016semi}} & - &\tabnum{\phantom{0}.877}{.036} & \tabnum{\phantom{0}.593}{.035} & \tabnum{\phantom{0}.845}{.051}& \tabnum{\phantom{0}.854}{.011} & \tabnum{\phantom{0}.772}{.041} & \tabnum{\phantom{0}.650}{.025} & 13.2

\\
 SchNet~{\small\citep{schutt2017schnet}}& - &\tabnum{\phantom{0}.847}{.024} & \tabnum{\phantom{0}.545}{.038} & \tabnum{\phantom{0}.717}{.042}& \tabnum{\phantom{0}.750}{.033} & \tabnum{\phantom{0}.767}{.025} & \tabnum{\phantom{0}.679}{.021} & 16.2
\\
 MPNN~{\small\citep{gilmer2017neural}}& - &\tabnum{\phantom{0}.913}{.041} & \tabnum{\phantom{0}.595}{.030} & \tabnum{\phantom{0}.879}{.054}& \tabnum{\phantom{0}.815}{.044} & \tabnum{\phantom{0}.808}{.024} & \tabnum{\phantom{0}.691}{.013} & 11.0
\\
 DMPNN~{\small\citep{yang2019analyzing}} & - &\tabnum{\phantom{0}.919}{.030} & \tabnum{\phantom{0}.632}{.023} & \tabnum{\phantom{0}.897}{.040}& \tabnum{\phantom{0}.852}{.053} & \cellcolor{yellowtwo}\tabnum{\phantom{0}.826}{.023} & \tabnum{\phantom{0}.718}{.011} & \phantom{0}5.7
\\
 MGCN~{\small\citep{lu2019molecular}}& - &\tabnum{\phantom{0}.850}{.064} & \tabnum{\phantom{0}.552}{.018} & \tabnum{\phantom{0}.634}{.042}& \tabnum{\phantom{0}.734}{.030} & \tabnum{\phantom{0}.707}{.016} & \tabnum{\phantom{0}.663}{.009} & 17.0
\\
 AttentiveFP~{\small\citep{xiong2019pushing}}& - &\tabnum{\phantom{0}.908}{.050} & \tabnum{\phantom{0}.605}{.060} & \cellcolor{yellowtwo}\tabnum{\phantom{0}.933}{.020}& \tabnum{\phantom{0}.863}{.015} & \tabnum{\phantom{0}.807}{.020} & \tabnum{\phantom{0}.579}{.001} & 10.0
\\

 CMPNN~{\small\citep{song2020communicative}} & - &\cellcolor{yellow}\besttabnum{\phantom{0}.940}{.009} & \tabnum{\phantom{0}.612}{.006} & \tabnum{\phantom{0}.931}{.003}& \tabnum{\phantom{0}.868}{.033} & \tabnum{\phantom{0}.805}{.017} & \tabnum{\phantom{0}.722}{.005} & \phantom{0}5.0
\\
 CoMPT~{\small\citep{chen2021learning}} & - &\tabnum{\phantom{0}.930}{.019} & \tabnum{\phantom{0}.605}{.011} & \tabnum{\phantom{0}.818}{.081}& \tabnum{\phantom{0}.851}{.043} & \tabnum{\phantom{0}.790}{.031} & \tabnum{\phantom{0}.716}{.010} & \phantom{0}9.5
\\

\midrule[0.5pt]
 N-GRAM~{\small\citep{liu2019n}}& \cmark &\tabnum{\phantom{0}.912}{.013} & \tabnum{\phantom{0}.632}{.005} & \tabnum{\phantom{0}.855}{.037}& \tabnum{\phantom{0}.876}{.035} & \tabnum{\phantom{0}.769}{.027} & - & \phantom{0}8.4
\\
 Hu. et.al~{\small\citep{hu2019strategies}} & \cmark &\tabnum{\phantom{0}.915}{.040} & \tabnum{\phantom{0}.614}{.006} & \tabnum{\phantom{0}.762}{.058}& \tabnum{\phantom{0}.851}{.027} & \tabnum{\phantom{0}.811}{.015} & \tabnum{\phantom{0}.714}{.019} & \phantom{0}9.2
\\
 MAT~{\small\citep{maziarka2020molecule}} & \cmark &\tabnum{\phantom{0}.922}{.035} & \tabnum{\phantom{0}.617}{.012} & \tabnum{\phantom{0}.853}{.079}& \tabnum{\phantom{0}.830}{.045} & \tabnum{\phantom{0}.810}{.015} & \tabnum{\phantom{0}.712}{.004} & \phantom{0}8.5
\\
 GROVER~{\small\citep{rong2020grover}} & \cmark &\tabnum{\phantom{0}.936}{.008} & \cellcolor{yellow}\besttabnum{\phantom{0}.656}{.006} & \tabnum{\phantom{0}.925}{.013}& \cellcolor{yellow}\besttabnum{\phantom{0}.878}{.016} & \tabnum{\phantom{0}.819}{.020} & \cellcolor{yellowtwo}\tabnum{\phantom{0}.723}{.010} & \cellcolor{yellowtwo}\phantom{0}2.3
\\
 Graphormer~{\small\citep{ying2021transformers}}& \cmark &\cellcolor{yellowtwo}\tabnum{\phantom{0}.938}{.032} & \tabnum{\phantom{0}.625}{.009} & \tabnum{\phantom{0}.913}{.056}& \tabnum{\phantom{0}.848}{.023} & \tabnum{\phantom{0}.801}{.013} & \tabnum{\phantom{0}.718}{.007} & \phantom{0}6.7
\\

 {GraphLoG~{\small\citep{xu2021self}}}& {\cmark} & {\tabnum{\phantom{0}.913}{.024}} & {\tabnum{\phantom{0}.595}{.039}} & {-} & {\tabnum{\phantom{0}.845}{.012}} & {\tabnum{\phantom{0}.773}{.010}} & {\tabnum{\phantom{0}.677}{.008}} & 13.0
\\
 {MPG*~{\small\citep{li2021effective}}}& {\cmark} & {\tabnum{\phantom{0}.922}{.039}} & {\tabnum{\phantom{0}.628}{.014}} & {-} & {\tabnum{\phantom{0}.864}{.028}} & {\tabnum{\phantom{0}.800}{.024}} & {\tabnum{\phantom{0}.712}{.009}} & {\phantom{0}7.4}
\\
 {GEM~{\small\citep{fang2022geometry}}}& {\cmark} & {\tabnum{\phantom{0}.921}{.026}} & {\tabnum{\phantom{0}.603}{.012}} & {-} & {\tabnum{\phantom{0}.872}{.036}} & {\tabnum{\phantom{0}.815}{.016}} & {\tabnum{\phantom{0}.720}{.010}} & {\phantom{0}6.6}
\\

\midrule[0.5pt]
 Ours & \cmark &\tabnum{\phantom{0}.934}{.018} & \cellcolor{yellowtwo}\tabnum{\phantom{0}.646}{.009} & \cellcolor{yellow}\besttabnum{\phantom{0}.935}{.014}& 
 \cellcolor{yellowtwo}\tabnum{\phantom{0}.877}{.032} & \cellcolor{yellow}\besttabnum{\phantom{0}.829}{.013} & \cellcolor{yellow}\besttabnum{\phantom{0}.730}{.005} & \cellcolor{yellow}\phantom{0}\textbf{1.8}
\\
\midrule[1.0pt]
\end{tabular}
}
\end{center}
\vspace{-1em}

\begin{centering}
\caption*{Regression Tasks}
\vspace{-0.1em}
\resizebox{\textwidth}{!}{
\begin{tabular}{lcccccccc}
\midrule[1.0pt]
 Method & Pre. & \hspace{0.12in}FreeSolv~$\downarrow$\hspace{0.12in} & \hspace{0.105in}ESOL~$\downarrow$\hspace{0.105in} & \hspace{0.105in}Lipo~$\downarrow$\hspace{0.105in} & \hspace{0.105in}QM7~$\downarrow$\hspace{0.105in} & \hspace{0.105in}QM8~$\downarrow$\hspace{0.105in} & Rank
\\ 
\midrule[0.5pt]
 TF\_Robust~{\small\citep{ramsundar2015massively}}& - &\tabnum{4.122}{.085} & \tabnum{1.722}{.038} & \tabnum{\phantom{0}.909}{.060}& \tabnum{120.6}{9.6}\phantom{\small0} & \tabnum{\phantom{0}.024}{.001} & 15.2
\\
 Weave~{\small\citep{kearnes2016molecular}}& - &\tabnum{2.398}{.250} & \tabnum{1.158}{.055} & \tabnum{\phantom{0}.813}{.042}& \tabnum{\phantom{0}94.7}{2.7}\phantom{\small0} & \tabnum{\phantom{0}.022}{.001} & 11.8
\\
 GraphConv~{\small\citep{kipf2016semi}} & - &\tabnum{2.900}{.135} & \tabnum{1.068}{.050} & \tabnum{\phantom{0}.712}{.049}& \tabnum{118.9}{20.2} & \tabnum{\phantom{0}.021}{.001} & 12.2

\\
 SchNet~{\small\citep{schutt2017schnet}}& - &\tabnum{3.215}{.755} & \tabnum{1.045}{.064} & \tabnum{\phantom{0}.909}{.098}& \tabnum{\phantom{0}74.2}{6.0}\phantom{\small0} & \tabnum{\phantom{0}.020}{.002} & 11.4
\\
 MPNN~{\small\citep{gilmer2017neural}}& - &\tabnum{2.185}{.952} & \tabnum{1.167}{.430} & \tabnum{\phantom{0}.672}{.051}& \tabnum{113.0}{17.2} & \tabnum{\phantom{0}.015}{.002} & 10.0
\\
 DMPNN~{\small\citep{yang2019analyzing}} & - &\tabnum{2.177}{.914} & \tabnum{\phantom{0}.980}{.258} & \tabnum{\phantom{0}.653}{.046}& \tabnum{105.8}{13.2} & \tabnum{.0143}{.002} & \phantom{0}7.8
\\
 MGCN~{\small\citep{lu2019molecular}}& - &\tabnum{3.349}{.097} & \tabnum{1.266}{.147} & \tabnum{1.113}{.041}& \phantom{0}\tabnum{77.6}{4.7}\phantom{\small0} & \tabnum{\phantom{0}.022}{.002} & 13.6
\\
 AttentiveFP~{\small\citep{xiong2019pushing}}& - &\tabnum{2.030}{.420} & \tabnum{\phantom{0}.853}{.060} & \tabnum{\phantom{0}.650}{.030}& \tabnum{126.7}{4.0}\phantom{\small0} & \tabnum{.0282}{.001} & \phantom{0}8.8
\\

 CMPNN~{\small\citep{song2020communicative}} & - &\tabnum{2.254}{.356} & \tabnum{\phantom{0}.841}{.090} & \tabnum{\phantom{0}.606}{.040}& \tabnum{\phantom{0}70.6}{2.7}\phantom{\small0} & \tabnum{.0136}{.001} & \phantom{0}4.8
\\
 CoMPT~{\small\citep{chen2021learning}} & - &\tabnum{2.125}{.590} & \tabnum{\phantom{0}.898}{.053} & \tabnum{\phantom{0}.632}{.038}& \tabnum{\phantom{0}65.3}{3.4}\phantom{\small0} & \tabnum{.0145}{.002} & \phantom{0}5.6
\\

\midrule[0.5pt]
 N-GRAM~{\small\citep{liu2019n}}& \cmark &\tabnum{2.512}{.190} & \tabnum{1.100}{.160} & \tabnum{\phantom{0}.876}{.033}& \tabnum{125.6}{1.5}\phantom{\small0} & \tabnum{.0320}{.003} & 14.0
\\
 MAT~{\small\citep{maziarka2020molecule}} & \cmark & \tabnum{2.116}{.152} & \tabnum{\phantom{0}.833}{.122} & \tabnum{\phantom{0}.668}{.025}& \tabnum{\phantom{0}93.6}{13.8} & \tabnum{.0178}{.002} & \phantom{0}6.6
\\
 GROVER~{\small\citep{rong2020grover}} & \cmark & \cellcolor{yellow}\besttabnum{1.592}{.072} & \tabnum{\phantom{0}.888}{.116} & \cellcolor{yellow}\besttabnum{\phantom{0}.563}{.030}& \tabnum{\phantom{0}72.5}{5.9}\phantom{\small0} & \tabnum{.0172}{.002} & \cellcolor{yellowtwo}\phantom{0}4.4
\\
 Graphormer~{\small\citep{ying2021transformers}}& \cmark & \tabnum{2.089}{.150}  & \cellcolor{yellowtwo}\tabnum{\phantom{0}.827}{.086} & \tabnum{\phantom{0}.674}{.035}& \tabnum{171.3}{10.7} & \tabnum{.0140}{.002} & \phantom{0}7.2
\\
 {MPG*~{\small\citep{li2021effective}}} & {\cmark} & {\tabnum{2.44}{.520}} & {\tabnum{\phantom{0}.926}{.159}} & {\tabnum{\phantom{0}.682}{.013}} & {-} & {-} & {10.7}
\\
 {GEM~{\small\citep{fang2022geometry}}} & {\cmark} & {\tabnum{2.21}{.374}} & {\tabnum{\phantom{0}.885}{.115}} & {\tabnum{\phantom{0}.617}{.023}} & {\cellcolor{yellow}\besttabnum{\phantom{0}58.9}{3.9}\phantom{\small0}} & {\cellcolor{yellowtwo}\tabnum{.0135}{.001}} & {\cellcolor{yellowtwo}\phantom{0}4.4}

\\
\midrule[0.5pt]
 Ours & \cmark & \cellcolor{yellowtwo}\tabnum{1.750}{.170} & \cellcolor{yellow}\besttabnum{\phantom{0}.822}{.073} & \cellcolor{yellowtwo}\tabnum{\phantom{0}.575}{.009}& \cellcolor{yellowtwo}\phantom{0}\tabnum{63.5}{3.4}\phantom{\small0} & \cellcolor{yellow}\besttabnum{.0134}{.002} & \cellcolor{yellow}\phantom{0}\textbf{1.6}
\\
\midrule[1.0pt]
\end{tabular}
}
\end{centering}

\vspace{-1em}
\end{table*}

\paragraph{Results}
\autoref{tab:results} shows the overall results of the baselines and our model on 11 MoleculeNet datasets. Our model achieved the best performance on five downstream tasks: ClinTox, Tox21, ToxCast, ESOL and QM8, and the second best performance on five downstream tasks: SIDER, BACE, FreeSolv, Lipo and QM7. We also computed the average rank for the classification and regression tasks, separately. Our model achieved the best average rank among all compared models. The result shows that our model generalizes well across different downstream tasks, which means local information aggregated from GNN can propagate globally as interacting with structural information in Transformer. We find that some substructures weight high attention to nodes consisting of the substructures, and CLS token used for prediction also focuses on the substructures from attention scores in self-attention.

\subsection{Ablation study}
We report ablation study to justify each component and flexibility of our model architecture. For the ablation study, we reduce the hyperparameter search space to only 6 learning rates \{1e-3, 5e-4, 1e-4, 5e-5, 1e-5, 5e-6\} to facilitate the comparison between different models. We report the test scores based on the hyperparameters from the best validation scores. 

\paragraph{Ablation on model components}
\begin{figure}[t!]
    \includegraphics[width=\linewidth]{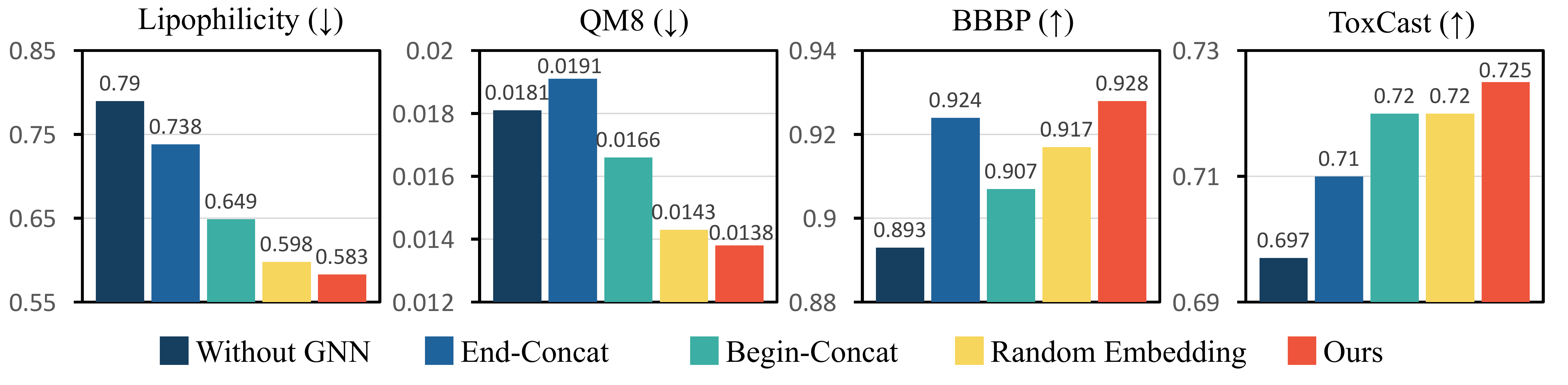}
    \caption{Ablation study. Four variations of our model (Without GNN, End-Concat, Begin-Concat, Random Embedding) on four downstream datasets. The results show the necessity of each component of our model.
    }
\label{fig:ablation_bargraph}
\vspace{-2em}

\end{figure}

\autoref{fig:ablation_bargraph} shows the performance comparison between four variations of our model on four downstream datasets. 
We first verify the performance of our model without GNN branch. %
To do that, we replace all cross-attention layers with self-attention layers and exclude GNN branch. 
Begin-Concat and End-Concat examine different ways of combining substructure and local features. 
Begin-Concat runs a vanilla-transformer encoder on top of a concatenated atom and substructure embeddings. 
End-Concat runs a vanilla-transformer encoder on top of atom embeddings and concatenates the substructure feature at the end to make the final prediction. 
Random Embedding examines the effect of substructural embeddings by replacing them with random learnable embeddings. %
\autoref{fig:ablation_bargraph} shows our model outperforms other variations on all datasets, which justifies the necessity of each component.

\paragraph{Different GNN branch}

\begin{figure}[t!]
    \centering
    \begin{minipage}[t]{0.33\linewidth}
        \includegraphics[width=\linewidth]{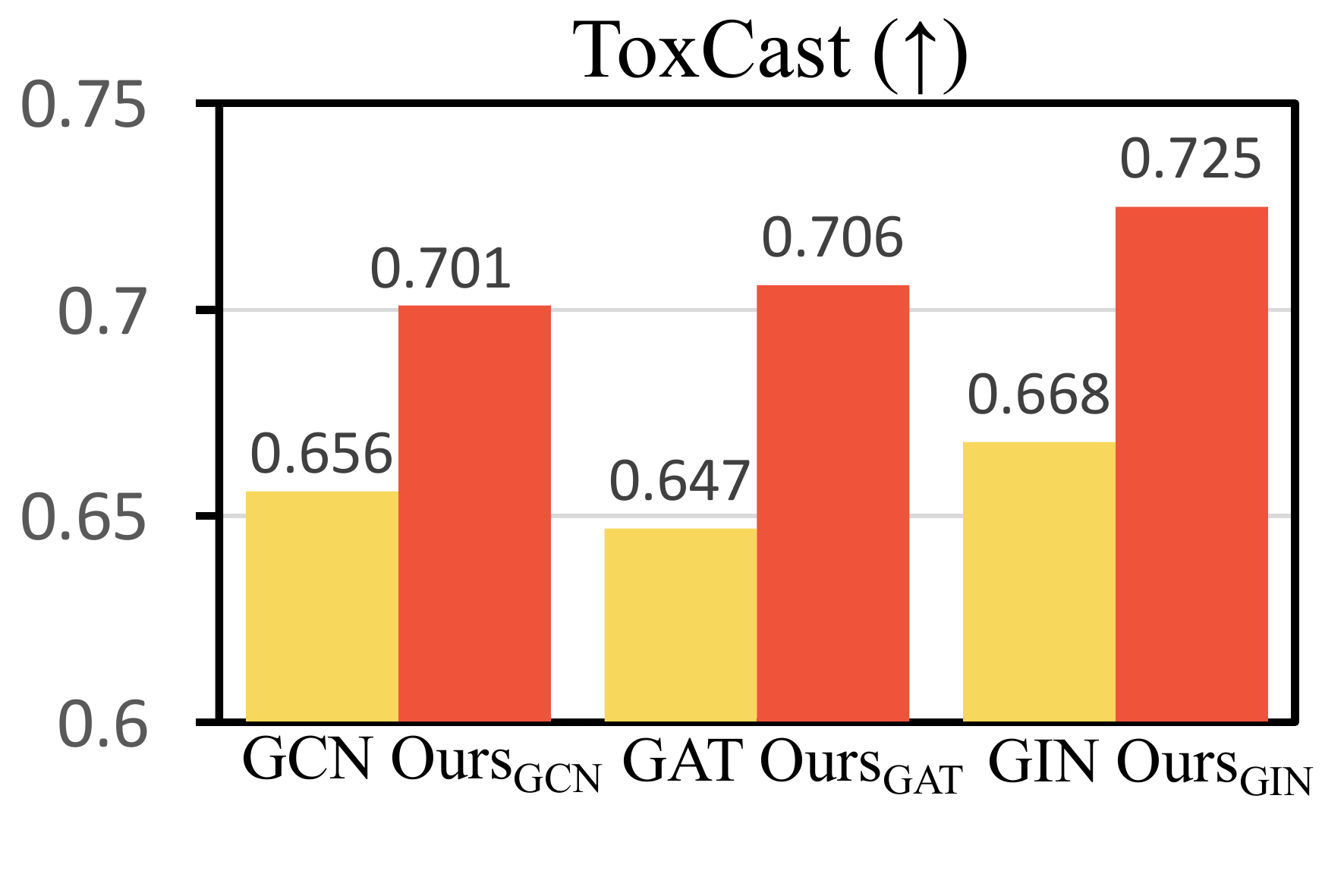}
        \caption{Comparison between standard three GNNs and our model on ToxCast dataset.}
        \label{fig:gnn_comp}
    \end{minipage}
    \hfill
    \begin{minipage}[t]{0.60\linewidth}
        \includegraphics[width=\linewidth]{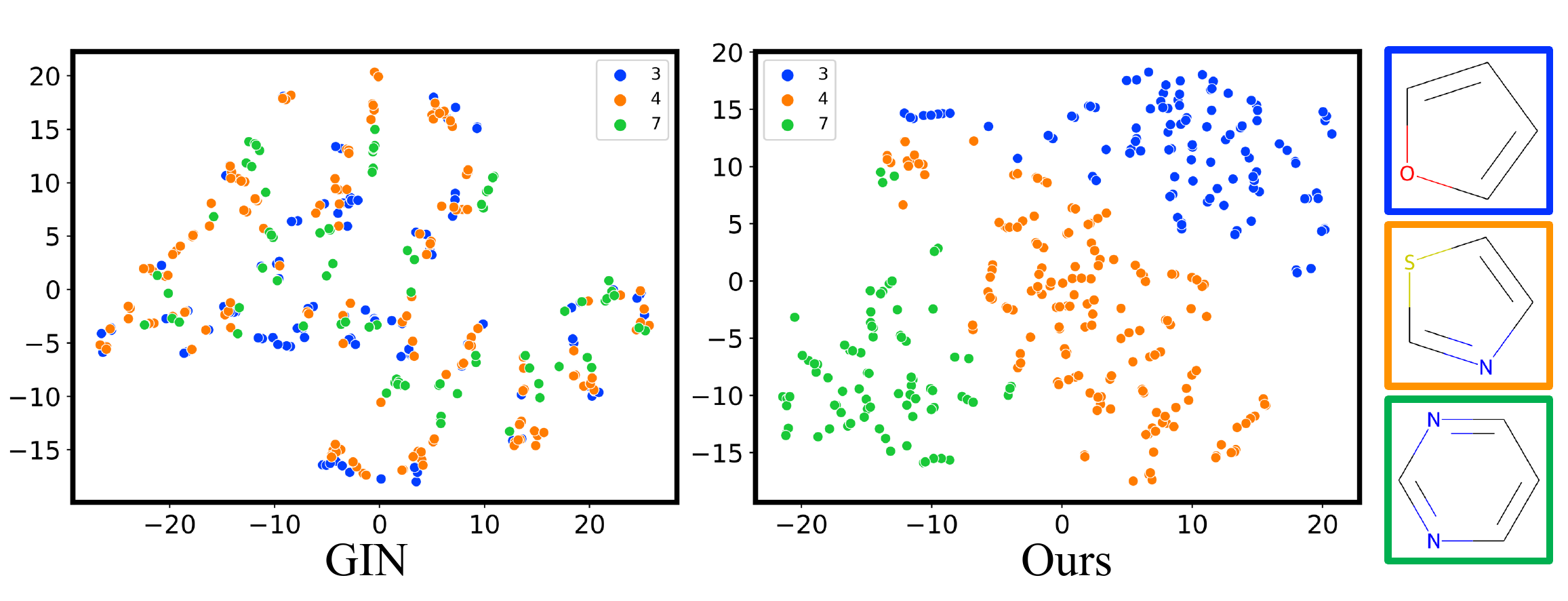}
        \caption{t-SNE embedding of molecules with different substructures.}
        \label{fig:tsne_embedding}
    \end{minipage}
\vspace{-1em}

\end{figure}

Our model can flexibly utilize other GNN architectures as our GNN branch. We test the changes in performance when our model design is applied with different GNN architectures. \autoref{fig:gnn_comp} shows the comparison between commonly-used GNNs, i.e., Graph Convolutional Network (GCN), Graph Attention Network (GAT)~\citep{velivckovic2017graph} and GIN, and our models with the GNN branch switched to each corresponding GNN on ToxCast dataset. There is a significant performance improvement when our model is adopted to each GNN model. Performance comparison when utilizing 3D-aware GNN is presented in \autoref{appendix:additional_ablation}. 

\newpage
\subsection{Analysis}

\paragraph{Effectiveness of substructures}
\autoref{fig:tsne_embedding} shows the t-SNE embeddings of molecules with different substructures. Our model discriminates molecules with similar substructures (aromatic rings with one or two different atoms) better than GIN. 

\paragraph{Attention map}

\begin{wrapfigure}{r}{0.5\textwidth}
\vspace{-2em}

  \centering
    \begin{subfigure}[b]{0.49\linewidth}
        \includegraphics[width=\linewidth]{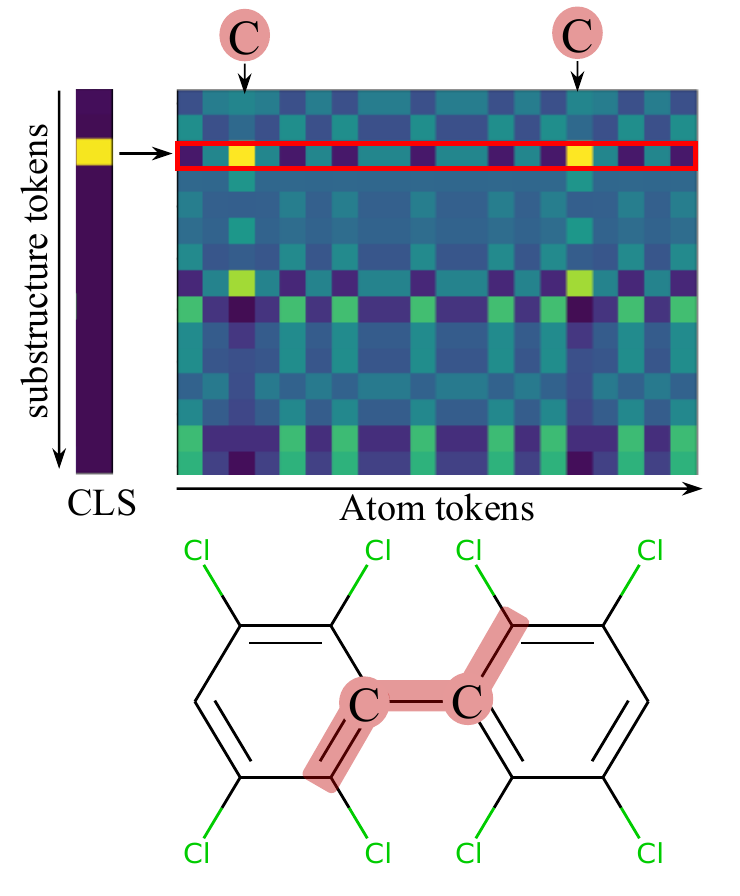}
        \caption{Ring-Chain-Ring bond }
        \label{fig:ring-chain-ring}
    \end{subfigure}
    \hfill
    \begin{subfigure}[b]{0.49\linewidth}
        \includegraphics[width=\linewidth]{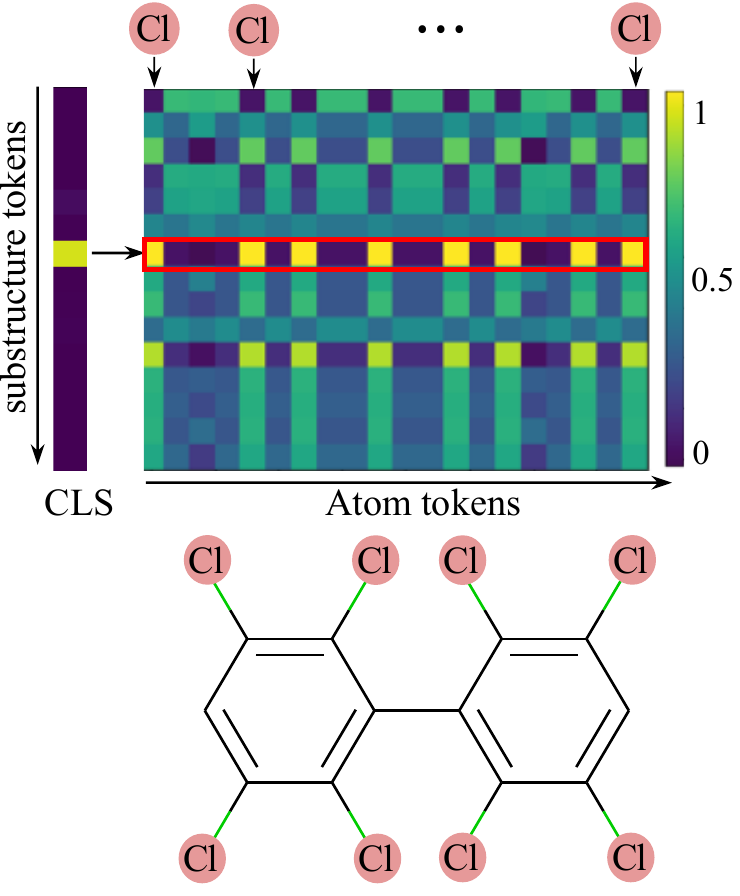}
        \caption{Chlorine atom}

        \label{fig:cl}
    \end{subfigure}
  \caption{Attention visualization. CLS token strongly attends to Ring-Chain-Ring bond and Cl substructures of the input molecule. The cross attention maps show the substructures capture the atoms related to the substructure.}
    \label{fig:attention_map}
\vspace{-2em}

\end{wrapfigure}

\autoref{fig:attention_map} shows the attention weights in a cross-attention and a self-attention layer. In the cross-attention layer, each row and column correspond to substructure and node in the given molecule, so we can interpret the attention score as the degree of focus between the substructures and nodes. We check where the CLS token gives more attention since it is used for prediction. We find out the CLS token gives strong attention to specific substructures. More interestingly, often a substructure gives more attention on the nodes that consists the substructure itself. Despite of not having any structural information between input substructures and nodes as input, our model can identify structurally related nodes in the cross attention layer, showcasing the ability of understanding molecular structure.

\paragraph{Computation time for graph features}
\begin{wrapfigure}{br}{0.5\textwidth}
\centering
\begin{minipage}{1.0\linewidth}
    \vspace{-5mm}
    \includegraphics[width=\linewidth]{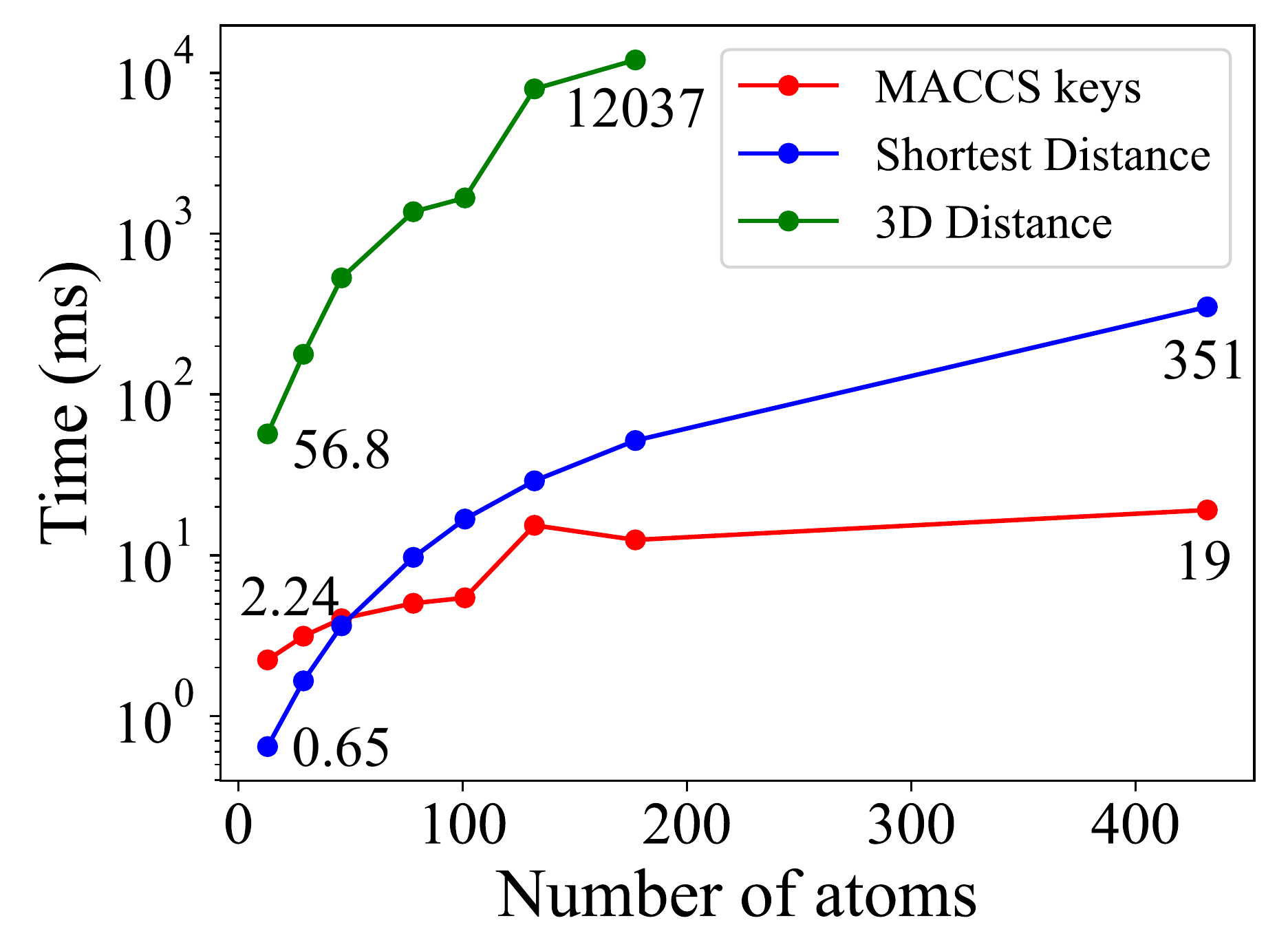}
    \vspace{-5mm}
    \caption{Time required to compute each feature per molecule with different number of atoms.%
    }
\label{fig:time_complex}
  \end{minipage}

\vspace{-1em}

\end{wrapfigure}

\autoref{fig:time_complex} shows the time required in millisecond to compute the MACCS keys, shortest path and 3D distance for each molecule with different number of atoms. As shown in the figure, the time required for 3D distance and the shortest path increases dramatically as the number of atoms increase, which make these features impossible to be applied in large molecules. However, identifying the MACCS key substructures does not depend on the input molecule's number of atoms.

\section{Conclusion}
\label{discussion}
In this paper, we propose a novel framework  that incorporates Transformer and GNN architecture for molecular representation learning. Our model takes advantages of the two architectures to aggregate substructure and local information. With the cross attention mechanism in fusing network, our model could achieve state-of-the-art performance on various molecular property prediction benchmarks. Overall, our work highlights the effectiveness of SACA for molecular representation learning. 

\paragraph{Limitation and future work} 
While we show the effectiveness of our model to predict molecular property, the proposed approach has not yet been validated with large-scale molecular graph such as proteins. Applying or modifying our model to proteins which contain 5,000 to 50,000 atoms would be an interesting direction for the future work.

\clearpage

\bibliographystyle{iclr2023_conference}
\bibliography{egbib}

\begin{thebibliography}{54}
\providecommand{\natexlab}[1]{#1}
\providecommand{\url}[1]{\texttt{#1}}
\expandafter\ifx\csname urlstyle\endcsname\relax
  \providecommand{\doi}[1]{doi: #1}\else
  \providecommand{\doi}{doi: \begingroup \urlstyle{rm}\Url}\fi

\bibitem[Alon \& Yahav(2020)Alon and Yahav]{alon2020bottleneck}
Uri Alon and Eran Yahav.
\newblock {On the Bottleneck of Graph Neural Networks and its Practical
  Implications}.
\newblock In \emph{Proceedings of the International Conference on Learning
  Representations (ICLR)}, 2020.

\bibitem[Bodnar et~al.(2021)Bodnar, Frasca, Otter, Wang, Lio, Montufar, and
  Bronstein]{bodnar2021weisfeiler}
Cristian Bodnar, Fabrizio Frasca, Nina Otter, Yuguang Wang, Pietro Lio, Guido~F
  Montufar, and Michael Bronstein.
\newblock Weisfeiler and lehman go cellular: Cw networks.
\newblock \emph{Advances in Neural Information Processing Systems},
  34:\penalty0 2625--2640, 2021.

\bibitem[Brown \& Martin(1996)Brown and Martin]{brown1996use}
Robert~D Brown and Yvonne~C Martin.
\newblock Use of structure- activity data to compare structure-based clustering
  methods and descriptors for use in compound selection.
\newblock \emph{Journal of Chemical Information and Computer Sciences},
  36\penalty0 (3):\penalty0 572--584, 1996.

\bibitem[Chen et~al.(2021)Chen, Zheng, Song, Rao, and Yang]{chen2021learning}
Jianwen Chen, Shuangjia Zheng, Ying Song, Jiahua Rao, and Yuedong Yang.
\newblock Learning attributed graph representations with communicative message
  passing transformer.
\newblock In \emph{IJCAI}, 2021.

\bibitem[Danel et~al.(2020)Danel, Spurek, Tabor, {\'S}mieja, Struski,
  S{\l}owik, and Maziarka]{danel2020spatial}
Tomasz Danel, Przemys{\l}aw Spurek, Jacek Tabor, Marek {\'S}mieja, {\L}ukasz
  Struski, Agnieszka S{\l}owik, and {\L}ukasz Maziarka.
\newblock Spatial graph convolutional networks.
\newblock In \emph{International Conference on Neural Information Processing},
  pp.\  668--675. Springer, 2020.

\bibitem[Debnath et~al.(1991)Debnath, Lopez~de Compadre, Debnath, Shusterman,
  and Hansch]{debnath1991structure}
Asim~Kumar Debnath, Rosa~L Lopez~de Compadre, Gargi Debnath, Alan~J Shusterman,
  and Corwin Hansch.
\newblock Structure-activity relationship of mutagenic aromatic and
  heteroaromatic nitro compounds. correlation with molecular orbital energies
  and hydrophobicity.
\newblock \emph{Journal of medicinal chemistry}, 34\penalty0 (2):\penalty0
  786--797, 1991.

\bibitem[Degen et~al.(2008)Degen, Wegscheid-Gerlach, Zaliani, and
  Rarey]{degen2008art}
J{\"o}rg Degen, Christof Wegscheid-Gerlach, Andrea Zaliani, and Matthias Rarey.
\newblock On the art of compiling and using'drug-like'chemical fragment spaces.
\newblock \emph{ChemMedChem: Chemistry Enabling Drug Discovery}, 3\penalty0
  (10):\penalty0 1503--1507, 2008.

\bibitem[Devlin et~al.(2019)Devlin, Chang, Lee, and
  Toutanova]{devlin-etal-2019-bert}
Jacob Devlin, Ming-Wei Chang, Kenton Lee, and Kristina Toutanova.
\newblock {BERT: Pre-training of Deep Bidirectional Transformers for Language
  Understanding}.
\newblock In \emph{Proceedings of the 2019 Conference of the North {A}merican
  Chapter of the Association for Computational Linguistics: Human Language
  Technologies, Volume 1 (Long and Short Papers)}, 2019.

\bibitem[Dosovitskiy et~al.(2020)Dosovitskiy, Beyer, Kolesnikov, Weissenborn,
  Zhai, Unterthiner, Dehghani, Minderer, Heigold, Gelly,
  et~al.]{dosovitskiy2020image}
Alexey Dosovitskiy, Lucas Beyer, Alexander Kolesnikov, Dirk Weissenborn,
  Xiaohua Zhai, Thomas Unterthiner, Mostafa Dehghani, Matthias Minderer, Georg
  Heigold, Sylvain Gelly, et~al.
\newblock An image is worth 16x16 words: Transformers for image recognition at
  scale.
\newblock In \emph{Proceedings of the International Conference on Learning
  Representations (ICLR)}, 2020.

\bibitem[Durant et~al.(2002)Durant, Leland, Henry, and
  Nourse]{durant2002reoptimization}
Joseph~L Durant, Burton~A Leland, Douglas~R Henry, and James~G Nourse.
\newblock {Reoptimization of MDL keys for use in drug discovery}.
\newblock \emph{Journal of chemical information and computer sciences}, 2002.

\bibitem[Eckert \& Bajorath(2007)Eckert and Bajorath]{eckert2007molecular}
Hanna Eckert and J{\"u}rgen Bajorath.
\newblock Molecular similarity analysis in virtual screening: foundations,
  limitations and novel approaches.
\newblock \emph{Drug discovery today}, 12\penalty0 (5-6):\penalty0 225--233,
  2007.

\bibitem[Fabian et~al.(2020)Fabian, Edlich, Gaspar, Segler, Meyers, Fiscato,
  and Ahmed]{fabian2020molecular}
Benedek Fabian, Thomas Edlich, H{\'e}l{\'e}na Gaspar, Marwin Segler, Joshua
  Meyers, Marco Fiscato, and Mohamed Ahmed.
\newblock Molecular representation learning with language models and
  domain-relevant auxiliary tasks.
\newblock \emph{arXiv preprint arXiv:2011.13230}, 2020.

\bibitem[Fang et~al.(2022)Fang, Liu, Lei, He, Zhang, Zhou, Wang, Wu, and
  Wang]{fang2022geometry}
Xiaomin Fang, Lihang Liu, Jieqiong Lei, Donglong He, Shanzhuo Zhang, Jingbo
  Zhou, Fan Wang, Hua Wu, and Haifeng Wang.
\newblock Geometry-enhanced molecular representation learning for property
  prediction.
\newblock \emph{Nature Machine Intelligence}, 4\penalty0 (2):\penalty0
  127--134, 2022.

\bibitem[Gaulton et~al.(2012)Gaulton, Bellis, Bento, Chambers, Davies, Hersey,
  Light, McGlinchey, Michalovich, Al-Lazikani, et~al.]{gaulton2012chembl}
Anna Gaulton, Louisa~J Bellis, A~Patricia Bento, Jon Chambers, Mark Davies,
  Anne Hersey, Yvonne Light, Shaun McGlinchey, David Michalovich, Bissan
  Al-Lazikani, et~al.
\newblock Chembl: a large-scale bioactivity database for drug discovery.
\newblock \emph{Nucleic acids research}, 2012.

\bibitem[Gilmer et~al.(2017)Gilmer, Schoenholz, Riley, Vinyals, and
  Dahl]{gilmer2017neural}
Justin Gilmer, Samuel~S Schoenholz, Patrick~F Riley, Oriol Vinyals, and
  George~E Dahl.
\newblock Neural message passing for quantum chemistry.
\newblock In \emph{International conference on machine learning}, pp.\
  1263--1272. PMLR, 2017.

\bibitem[Hu et~al.(2019)Hu, Liu, Gomes, Zitnik, Liang, Pande, and
  Leskovec]{hu2019strategies}
Weihua Hu, Bowen Liu, Joseph Gomes, Marinka Zitnik, Percy Liang, Vijay Pande,
  and Jure Leskovec.
\newblock {Strategies for Pre-training Graph Neural Networks}.
\newblock In \emph{Proceedings of the International Conference on Learning
  Representations (ICLR)}, 2019.

\bibitem[Hu et~al.(2020)Hu, Fey, Zitnik, Dong, Ren, Liu, Catasta, and
  Leskovec]{hu2020ogb}
Weihua Hu, Matthias Fey, Marinka Zitnik, Yuxiao Dong, Hongyu Ren, Bowen Liu,
  Michele Catasta, and Jure Leskovec.
\newblock Open graph benchmark: Datasets for machine learning on graphs.
\newblock \emph{arXiv preprint arXiv:2005.00687}, 2020.

\bibitem[Hughes et~al.(2011)Hughes, Rees, Kalindjian, and
  Philpott]{hughes2011principles}
James~P Hughes, Stephen Rees, S~Barrett Kalindjian, and Karen~L Philpott.
\newblock Principles of early drug discovery.
\newblock \emph{British journal of pharmacology}, 162\penalty0 (6):\penalty0
  1239--1249, 2011.

\bibitem[Jin et~al.(2018)Jin, Barzilay, and Jaakkola]{jin2018junction}
Wengong Jin, Regina Barzilay, and Tommi Jaakkola.
\newblock Junction tree variational autoencoder for molecular graph generation.
\newblock In \emph{Proceedings of the International Conference on Learning
  Representations (ICLR)}, 2018.

\bibitem[Kearnes et~al.(2016)Kearnes, McCloskey, Berndl, Pande, and
  Riley]{kearnes2016molecular}
Steven Kearnes, Kevin McCloskey, Marc Berndl, Vijay Pande, and Patrick Riley.
\newblock Molecular graph convolutions: moving beyond fingerprints.
\newblock \emph{Journal of computer-aided molecular design}, pp.\  595--608,
  2016.

\bibitem[Kim et~al.(2016)Kim, Thiessen, Bolton, Chen, Fu, Gindulyte, Han, He,
  He, Shoemaker, et~al.]{kim2016pubchem}
Sunghwan Kim, Paul~A Thiessen, Evan~E Bolton, Jie Chen, Gang Fu, Asta
  Gindulyte, Lianyi Han, Jane He, Siqian He, Benjamin~A Shoemaker, et~al.
\newblock {PubChem substance and compound databases}.
\newblock \emph{Nucleic acids research}, 2016.

\bibitem[Kipf \& Welling(2017)Kipf and Welling]{kipf2016semi}
Thomas~N Kipf and Max Welling.
\newblock Semi-supervised classification with graph convolutional networks.
\newblock \emph{Proceedings of the International Conference on Learning
  Representations (ICLR)}, 2017.

\bibitem[Landrum(2016)]{Landrum2016RDKit2016_09_4}
Greg Landrum.
\newblock Rdkit: Open-source cheminformatics software.
\newblock 2016.
\newblock URL
  \url{https://github.com/rdkit/rdkit/releases/tag/Release_2016_09_4}.

\bibitem[Li et~al.(2019)Li, Koh, Reker, Brown, Wang, Lee, Liow, Dai, Fan, Chen,
  et~al.]{li2019predicting}
Li~Li, Ching~Chiek Koh, Daniel Reker, JB~Brown, Haishuai Wang, Nicholas~Keone
  Lee, Hien-haw Liow, Hao Dai, Huai-Meng Fan, Luonan Chen, et~al.
\newblock Predicting protein-ligand interactions based on bow-pharmacological
  space and bayesian additive regression trees.
\newblock \emph{Scientific reports}, 9\penalty0 (1):\penalty0 1--12, 2019.

\bibitem[Li et~al.(2021)Li, Wang, Qiao, Chen, Yu, Yao, Gao, Xie, and
  Song]{li2021effective}
Pengyong Li, Jun Wang, Yixuan Qiao, Hao Chen, Yihuan Yu, Xiaojun Yao, Peng Gao,
  Guotong Xie, and Sen Song.
\newblock An effective self-supervised framework for learning expressive
  molecular global representations to drug discovery.
\newblock \emph{Briefings in Bioinformatics}, 22\penalty0 (6):\penalty0
  bbab109, 2021.

\bibitem[Li et~al.(2018)Li, Han, and Wu]{li2018deeper}
Qimai Li, Zhichao Han, and Xiao-Ming Wu.
\newblock {Deeper insights into graph convolutional networks for
  semi-supervised learning}.
\newblock In \emph{Thirty-Second AAAI conference on artificial intelligence},
  2018.

\bibitem[Liu et~al.(2019)Liu, Demirel, and Liang]{liu2019n}
Shengchao Liu, Mehmet~F Demirel, and Yingyu Liang.
\newblock N-gram graph: Simple unsupervised representation for graphs, with
  applications to molecules.
\newblock \emph{Advances in neural information processing systems}, 32, 2019.

\bibitem[Loshchilov \& Hutter(2018)Loshchilov and
  Hutter]{loshchilov2018decoupled}
Ilya Loshchilov and Frank Hutter.
\newblock Decoupled weight decay regularization.
\newblock In \emph{International Conference on Learning Representations}, 2018.

\bibitem[Lu et~al.(2019)Lu, Liu, Wang, Huang, Lin, and He]{lu2019molecular}
Chengqiang Lu, Qi~Liu, Chao Wang, Zhenya Huang, Peize Lin, and Lixin He.
\newblock Molecular property prediction: A multilevel quantum interactions
  modeling perspective.
\newblock In \emph{Proceedings of the AAAI Conference on Artificial
  Intelligence}, volume~33, pp.\  1052--1060, 2019.

\bibitem[Maziarka et~al.(2020)Maziarka, Danel, Mucha, Rataj, Tabor, and
  Jastrz{\k{e}}bski]{maziarka2020molecule}
{\L}ukasz Maziarka, Tomasz Danel, S{\l}awomir Mucha, Krzysztof Rataj, Jacek
  Tabor, and Stanis{\l}aw Jastrz{\k{e}}bski.
\newblock {Molecule attention transformer}.
\newblock \emph{arXiv preprint arXiv:2002.08264}, 2020.

\bibitem[Maziarka et~al.(2021)Maziarka, Majchrowski, Danel, Gai{\'n}ski, Tabor,
  Podolak, Morkisz, and Jastrz{\k{e}}bski]{maziarka2021relative}
{\L}ukasz Maziarka, Dawid Majchrowski, Tomasz Danel, Piotr Gai{\'n}ski, Jacek
  Tabor, Igor Podolak, Pawe{\l} Morkisz, and Stanis{\l}aw Jastrz{\k{e}}bski.
\newblock {Relative Molecule Self-Attention Transformer}.
\newblock \emph{arXiv preprint arXiv:2110.05841}, 2021.

\bibitem[Mohs \& Greig(2017)Mohs and Greig]{mohs2017drug}
Richard~C Mohs and Nigel~H Greig.
\newblock Drug discovery and development: Role of basic biological research.
\newblock \emph{Alzheimer's \& Dementia: Translational Research \& Clinical
  Interventions}, 3\penalty0 (4):\penalty0 651--657, 2017.

\bibitem[Murray \& Rees(2009)Murray and Rees]{murray2009rise}
Christopher~W Murray and David~C Rees.
\newblock The rise of fragment-based drug discovery.
\newblock \emph{Nature chemistry}, 2009.

\bibitem[Ramsundar et~al.(2015)Ramsundar, Kearnes, Riley, Webster, Konerding,
  and Pande]{ramsundar2015massively}
Bharath Ramsundar, Steven Kearnes, Patrick Riley, Dale Webster, David
  Konerding, and Vijay Pande.
\newblock Massively multitask networks for drug discovery.
\newblock \emph{arXiv preprint arXiv:1502.02072}, 2015.

\bibitem[Rogers \& Hahn(2010)Rogers and Hahn]{rogers2010extended}
David Rogers and Mathew Hahn.
\newblock Extended-connectivity fingerprints.
\newblock \emph{Journal of chemical information and modeling}, 2010.

\bibitem[Rong et~al.(2020)Rong, Bian, Xu, Xie, Wei, Huang, and
  Huang]{rong2020grover}
Yu~Rong, Yatao Bian, Tingyang Xu, Weiyang Xie, Ying Wei, Wenbing Huang, and
  Junzhou Huang.
\newblock {GROVER: Self-supervised Message Passing Transformer on Large-scale
  Molecular Data}.
\newblock In \emph{Advances in Neural Information Processing Systems
  (NeurIPS)}, 2020.

\bibitem[Sch{\"u}tt et~al.(2017)Sch{\"u}tt, Kindermans, Sauceda~Felix, Chmiela,
  Tkatchenko, and M{\"u}ller]{schutt2017schnet}
Kristof Sch{\"u}tt, Pieter-Jan Kindermans, Huziel~Enoc Sauceda~Felix, Stefan
  Chmiela, Alexandre Tkatchenko, and Klaus-Robert M{\"u}ller.
\newblock Schnet: A continuous-filter convolutional neural network for modeling
  quantum interactions.
\newblock \emph{Advances in neural information processing systems}, 30, 2017.

\bibitem[Song et~al.(2020)Song, Zheng, Niu, Fu, Lu, and
  Yang]{song2020communicative}
Ying Song, Shuangjia Zheng, Zhangming Niu, Zhang-Hua Fu, Yutong Lu, and Yuedong
  Yang.
\newblock Communicative representation learning on attributed molecular graphs.
\newblock In \emph{IJCAI}, 2020.

\bibitem[Vaswani et~al.(2017)Vaswani, Shazeer, Parmar, Uszkoreit, Jones, Gomez,
  Kaiser, and Polosukhin]{vaswani2017attention}
Ashish Vaswani, Noam Shazeer, Niki Parmar, Jakob Uszkoreit, Llion Jones,
  Aidan~N Gomez, {\L}ukasz Kaiser, and Illia Polosukhin.
\newblock {Attention is all you need}.
\newblock In \emph{Advances in Neural Information Processing Systems
  (NeurIPS)}, 2017.

\bibitem[Veli{\v{c}}kovi{\'c} et~al.(2018)Veli{\v{c}}kovi{\'c}, Cucurull,
  Casanova, Romero, Lio, and Bengio]{velivckovic2017graph}
Petar Veli{\v{c}}kovi{\'c}, Guillem Cucurull, Arantxa Casanova, Adriana Romero,
  Pietro Lio, and Yoshua Bengio.
\newblock Graph attention networks.
\newblock \emph{Proceedings of the International Conference on Learning
  Representations (ICLR)}, 2018.

\bibitem[Weininger(1988)]{weininger1988smiles}
David Weininger.
\newblock Smiles, a chemical language and information system. 1. introduction
  to methodology and encoding rules.
\newblock \emph{Journal of chemical information and computer sciences}, 1988.

\bibitem[Weisfeiler \& Leman(1968)Weisfeiler and
  Leman]{weisfeiler1968reduction}
Boris Weisfeiler and Andrei Leman.
\newblock The reduction of a graph to canonical form and the algebra which
  appears therein.
\newblock \emph{NTI, Series}, 2\penalty0 (9):\penalty0 12--16, 1968.

\bibitem[Willett et~al.(1998)Willett, Barnard, and Downs]{willett1998chemical}
Peter Willett, John~M Barnard, and Geoffrey~M Downs.
\newblock Chemical similarity searching.
\newblock \emph{Journal of chemical information and computer sciences},
  38\penalty0 (6):\penalty0 983--996, 1998.

\bibitem[Wu et~al.(2018)Wu, Ramsundar, Feinberg, Gomes, Geniesse, Pappu,
  Leswing, and Pande]{wu2018moleculenet}
Zhenqin Wu, Bharath Ramsundar, Evan~N Feinberg, Joseph Gomes, Caleb Geniesse,
  Aneesh~S Pappu, Karl Leswing, and Vijay Pande.
\newblock Moleculenet: a benchmark for molecular machine learning.
\newblock \emph{Chemical science}, 2018.

\bibitem[Xiong et~al.(2019)Xiong, Wang, Liu, Zhong, Wan, Li, Li, Luo, Chen,
  Jiang, et~al.]{xiong2019pushing}
Zhaoping Xiong, Dingyan Wang, Xiaohong Liu, Feisheng Zhong, Xiaozhe Wan, Xutong
  Li, Zhaojun Li, Xiaomin Luo, Kaixian Chen, Hualiang Jiang, et~al.
\newblock Pushing the boundaries of molecular representation for drug discovery
  with the graph attention mechanism.
\newblock \emph{Journal of medicinal chemistry}, 63\penalty0 (16):\penalty0
  8749--8760, 2019.

\bibitem[Xu et~al.(2018)Xu, Li, Tian, Sonobe, Kawarabayashi, and
  Jegelka]{xu2018representation}
Keyulu Xu, Chengtao Li, Yonglong Tian, Tomohiro Sonobe, Ken-ichi Kawarabayashi,
  and Stefanie Jegelka.
\newblock Representation learning on graphs with jumping knowledge networks.
\newblock In \emph{International Conference on Machine Learning}, pp.\
  5453--5462. PMLR, 2018.

\bibitem[Xu et~al.(2019)Xu, Hu, Leskovec, and Jegelka]{xu2018powerful}
Keyulu Xu, Weihua Hu, Jure Leskovec, and Stefanie Jegelka.
\newblock How powerful are graph neural networks?
\newblock \emph{Proceedings of the International Conference on Learning
  Representations (ICLR)}, 2019.

\bibitem[Xu et~al.(2021)Xu, Wang, Ni, Guo, and Tang]{xu2021self}
Minghao Xu, Hang Wang, Bingbing Ni, Hongyu Guo, and Jian Tang.
\newblock Self-supervised graph-level representation learning with local and
  global structure.
\newblock In \emph{International Conference on Machine Learning}, pp.\
  11548--11558. PMLR, 2021.

\bibitem[Yang et~al.(2021)Yang, Liu, Xiao, Li, Lian, Agrawal, Singh, Sun, and
  Xie]{yang2021graphformers}
Junhan Yang, Zheng Liu, Shitao Xiao, Chaozhuo Li, Defu Lian, Sanjay Agrawal,
  Amit Singh, Guangzhong Sun, and Xing Xie.
\newblock Graphformers: Gnn-nested transformers for representation learning on
  textual graph.
\newblock \emph{Advances in Neural Information Processing Systems},
  34:\penalty0 28798--28810, 2021.

\bibitem[Yang et~al.(2019)Yang, Swanson, Jin, Coley, Eiden, Gao, Guzman-Perez,
  Hopper, Kelley, Mathea, et~al.]{yang2019analyzing}
Kevin Yang, Kyle Swanson, Wengong Jin, Connor Coley, Philipp Eiden, Hua Gao,
  Angel Guzman-Perez, Timothy Hopper, Brian Kelley, Miriam Mathea, et~al.
\newblock Analyzing learned molecular representations for property prediction.
\newblock \emph{Journal of chemical information and modeling}, 2019.

\bibitem[Ying et~al.(2021)Ying, Cai, Luo, Zheng, Ke, He, Shen, and
  Liu]{ying2021transformers}
Chengxuan Ying, Tianle Cai, Shengjie Luo, Shuxin Zheng, Guolin Ke, Di~He,
  Yanming Shen, and Tie-Yan Liu.
\newblock {Do Transformers Really Perform Bad for Graph Representation?}
\newblock In \emph{Advances in Neural Information Processing Systems
  (NeurIPS)}, 2021.

\bibitem[Zhang et~al.(2021)Zhang, Liu, Wang, Lu, and Lee]{zhang2021motif}
Zaixi Zhang, Qi~Liu, Hao Wang, Chengqiang Lu, and Chee-Kong Lee.
\newblock Motif-based graph self-supervised learning for molecular property
  prediction.
\newblock \emph{Advances in Neural Information Processing Systems},
  34:\penalty0 15870--15882, 2021.

\bibitem[Zhu et~al.(2021{\natexlab{a}})Zhu, Xia, Qin, Zhou, Li, and
  Liu]{zhu2021dual}
Jinhua Zhu, Yingce Xia, Tao Qin, Wengang Zhou, Houqiang Li, and Tie-Yan Liu.
\newblock Dual-view molecule pre-training.
\newblock \emph{arXiv preprint arXiv:2106.10234}, 2021{\natexlab{a}}.

\bibitem[Zhu et~al.(2021{\natexlab{b}})Zhu, Xu, Shen, Ji, Gao, and
  Shen]{zhu2021posegtac}
Yiran Zhu, Xing Xu, Fumin Shen, Yanli Ji, Lianli Gao, and Heng~Tao Shen.
\newblock Posegtac: Graph transformer encoder-decoder with atrous convolution
  for 3d human pose estimation.
\newblock In \emph{IJCAI}, pp.\  1359--1365, 2021{\natexlab{b}}.

\end{thebibliography}

\clearpage
\appendix
\section{Node and Edge Features}
\label{appendix: features}
In this section, we present the node and edge features of molecules used for GNN branch. 
We used OGB package~\citep{hu2020ogb} to convert SMILES strings~\citep{weininger1988smiles} to molecular graphs. The molecular graphs are encoded through GIN~\citep{xu2018powerful} and injected into the Transformer branch by the cross-attention. The molecular graphs have the following node and edge features. 

\paragraph{Node features.}
Each node has the following 9 dimensional features as shown in \autoref{tab:node}.

\begin{table*}[ht]
\begin{center}
\caption{\label{tab:node}Node Features.}
\vskip 0.1in
\begin{tabular}{lcc}
\toprule
Index & Description & Range 
\\ 
\midrule
0 & Atomic num    & [1, 118], other
\\
1 & Chirality    & unspecified, tetrahedral cw, tetrahedral ccw, other
\\
2 & Degree    & [0, 10], other
\\
3 & Formal Charge    & [-5, 5], other
\\
4 & Num Hydrogen    & [0, 8], other
\\
5 & Num Radical Electron    & [0, 4], other
\\
6 & Hybridization    & SP, SP2, SP3, SP3D, SP3D2, other
\\
7 & Is Aromatic    & False, True
\\
8 & Is in Ring    & False, True
\\
\bottomrule
\end{tabular}
\end{center}
\end{table*}

\paragraph{Edge features.}
Each edge has the following 3 dimensional features as shown in \autoref{tab:edge}.
\begin{table*}[ht]
\begin{center}
\caption{\label{tab:edge}Edge Features.}
\vskip 0.1in
\begin{tabular}{lcc}
\toprule
Index & Description & Range 
\\ 
\midrule
0 & Bond Type    & single, double, triple, aromatic, other
\\
1 & Bond Stereo    & stereonone, stereoz, stereoe, stereocis, stereotrans, stereoany
\\
2 & Is Conjugated    & False, True
\\
\bottomrule
\end{tabular}
\end{center}
\end{table*}

\newpage
\section{Details of Downstream Datasets}
\label{appendix:downstream_dataset}
We used 11 binary graph classification and regression datasets: BBBP, SIDER, ClinTox, BACE, Tox21, ToxCast, FreeSolv, ESOL, Lipophilicity, QM7 and QM8 from Moleculenet~\citep{wu2018moleculenet}. The details of each dataset are shown in \autoref{tab:dataset details}. Through the various datasets, we can test the generalization ability of our pretrained model.

\begin{table*}[ht]
\begin{center}
\caption{Detailed description for each downstream dataset.}
\label{tab:dataset details}
\resizebox{\textwidth}{!}{
\begin{tabular}{ll}
\toprule
Dataset & Description 
\\ 
\midrule
BBBP & Binary classification task to predict a molecule's blood-brain barrier penetration ability
\\
SIDER & Marketed drugs with its adverse drug reactions
\\
ClinTox & Qualitative data of drugs approved by the FDA and those that have failed clinical trials for toxicity reasons
\\
BACE & Binary classification task to predict a molecule's binding result for a set of inhibitors of human $\beta$-secretase 1
\\
Tox21 & Qualitative toxicity measurements on 12 biological targets
\\
ToxCast & Toxicology data for a large library of compounds based on in vitro high-throughput screening, including experiments on over 600 tasks 
\\
FreeSolv & Regression task to predict hydration free energy of small molecules in water
\\
ESOL & Regression task to predict water solubility in terms of log solubility in mols per litre
\\
Lipophilicity & Experimental results of octanol/water distribution coefficient
\\
QM7 & A subset of GDB-13 composed of all molecules of up to 23 atoms (including 7 heavy atoms C, N, O, and S), totalling 7165 molecules
\\
QM8 & Computer-generated quantum mechanical properties
\\
\bottomrule
\end{tabular}
}
\end{center}
\end{table*}

\section{Molecular Structural Keys}

In this section, we explain the details of molecular substructures that we used for our model, i.e., Molecular ACCess System (MACCS) keys~\citep{durant2002reoptimization}. We extract MACCS keys for each molecule using RDKit package~\citep{Landrum2016RDKit2016_09_4}. MACCS keys have 166-dimensional features where each binary label indicates the presence of a particular substructure in the given molecule. For example, the $139^{th}$ index of MACCS keys indicates the presence of the Hydroxy group (-OH), and the $162^{nd}$ index indicates the presence of the aromatic ring in a molecule. The full list of 166 MACCS keys can be found in the document \footnote{\url{https://github.com/rdkit/rdkit/blob/master/rdkit/Chem/MACCSkeys.py}}. 

\begin{figure}[h]
    \includegraphics[width=\linewidth]{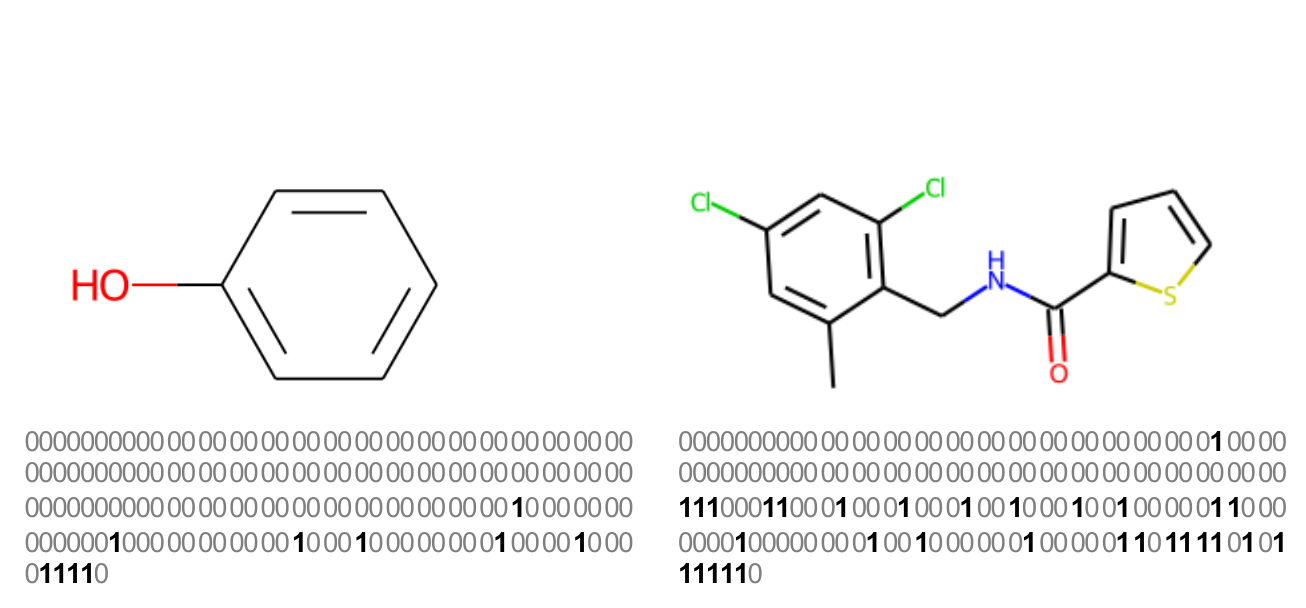}
    \caption{MACCS keys encoding.
    }
\label{fig:maccs_eg}
\end{figure}
\autoref{fig:maccs_eg} shows examples of how MACCS keys encode substructure information into a binary bit string. The canonical SMILES representations of the left and right molecules are \textsf{C1=CC=C(C=C1)O} and \textsf{Cc1cc(Cl)cc(Cl)c1CNC(=O)c1cccs1}, respectively. 

\autoref{tab:maccs_key} shows the MACCS keys statistics for pretraining and downstream datasets. It includes the average number of MACCS keys that molecules in each dataset contain and the standard deviation. Also, the minimum and the maximum number of MACCS keys for each dataset are reported. 

\begin{table*}[ht]
\begin{center}
\caption{MACCS keys statistics for pretraining and downstream dataset.}
\label{tab:maccs_key}
\vskip 0.1in
\begin{tabular}{lcccc}
\toprule
Dataset & Average & S.D & Min & Max  
\\ 
\midrule
Pretraining dataset & 51.99 & 13.44 & 1 & 106
\\
BBBP & 46.03 & 14.48 & 2 & 96
\\
SIDER & 46.61 & 17.93 & 1 & 105
\\
ClinTox & 46.13 & 16.54 & 2 & 92
\\
BACE & 61.04 & 12.53 & 21 & 93
\\
Tox21 & 32.73 & 16.63 & 2 & 99
\\
ToxCast & 33.50 & 17.30 & 2 & 101
\\
FreeSolv & 15.26 & 9.78 & 1 & 62
\\
ESOL & 23.63 & 15.52 & 1 & 76
\\
Lipophilicity & 51.61 & 14.26 & 7 & 93
\\
QM7 & 18.75 & 7.80 & 1 & 50
\\
QM8 & 23.00 & 8.10 & 1 & 48
\\
\bottomrule
\end{tabular}
\end{center}
\end{table*}

\section{Experimental Setting}
\label{appendix:experimental setting}
In this section, we explain further details of our experimental setting.

\paragraph{Pretraining.}
The default hyperparameters for pretraining are listed in \autoref{tab:model_config}. All attention layers have 16 attention heads and 768 hidden dimension. Our model has 41M parameters.
\begin{table*}[ht]
\begin{center}
\caption{Model configurations and hyperparameters for pretraining.}
\label{tab:model_config}
\vskip 0.1in
\begin{tabular}{lcc}
\toprule
 & Parameter 
\\ 
\midrule
M & 4
\\
N & 3
\\
Model hidden dimension & 768
\\
FFN inner-layer dimension & 768
\\
\# of attention heads & 16
\\
Learning rate & 0.0001
\\
Epoch & 10
\\
Dropout & 0.1
\\
Batch size & 32
\\
\bottomrule
\end{tabular}
\end{center}
\end{table*}

\paragraph{Hyperparameter Search Range.}
For each downstream task, we search for the best hyperparameter combinations. We perform the Bayesian optimization over the validation set and use the hyperparameters for the best validation score to report the test score. The hyperparameter range that we searched over is shown in \autoref{tab:hyperparameter}.

\begin{table*}[ht]
\begin{center}
\caption{Finetuning hyperparameter search range.}
\label{tab:hyperparameter}
\vskip 0.1in
\begin{tabular}{lll}
\toprule
Hyperparameter & Description & Range 
\\ 
\midrule
Learninng rate &  The learning rate & 0.000001 $\sim$ 0.001
\\
\# of MLP layers & The number of last MLP layers & 1, 2, 3
\\
Dropout & Dropout ratio & 0.0 $\sim$ 0.5
\\
Weight decay & Weight decay  & 0.0, 0.001, 0.0001, 0.00001
\\
\bottomrule
\end{tabular}
\end{center}
\end{table*}

\paragraph{Pretraining Task.}
To pretrain our network, we extract the 200 real-valued descriptors for each molecule using RDKit package~\citep{Landrum2016RDKit2016_09_4}. This task is proposed by~\citep{fabian2020molecular}. Through this task, we can make the model to learn the physicochemical properties of the input molecules.

\section{Parameters and space complexity of different models}

We present the number of parameters of transformer-based molecular representation learning models in \autoref{tab:param_complex}. Except for CoMPT, our model has a comparable or lower number of parameters than other transformer-based models. 

We also present the space complexity in \autoref{tab:param_complex}. The space complexity for our attention computation is linear to the number of atoms (i.e., $\mathcal{O}(n)$ where $n$ is the number of atoms). %
On the other hand, the space complexity of other transformer-based molecule representation learning models is quadratic to the number of atoms (i.e., $\mathcal{O}(n^2)$).

\begin{table*}[h]
\begin{center}

\caption{Parameter and complexity comparison between different models.}
\label{tab:param_complex}
\vskip 0.1in
\begin{tabular}{lcc}
\toprule
 & Parameters & Space Complexity for Attention    
\\
\midrule
GROVER~\citep{rong2020grover} & 48M & $\mathcal{O}(n^2)$
\\

Graphormer~\citep{ying2021transformers} & 47M	 & $\mathcal{O}(n^2)$
\\

MAT~\citep{maziarka2020molecule} & 42M & $\mathcal{O}(n^2)$
\\

CoMPT~\citep{chen2021learning} &  2.7M	 & $\mathcal{O}(n^2)$
\\
Ours  & 41M	 & $\mathcal{O}(n)$

\\

\bottomrule
\end{tabular}

\end{center}
\end{table*}

\section{Additional Ablation Study}
\label{appendix:additional_ablation}

\begin{wrapfigure}{r}{0.3\textwidth}
\centering
\begin{minipage}{1.0\linewidth}
    \includegraphics[width=\linewidth]{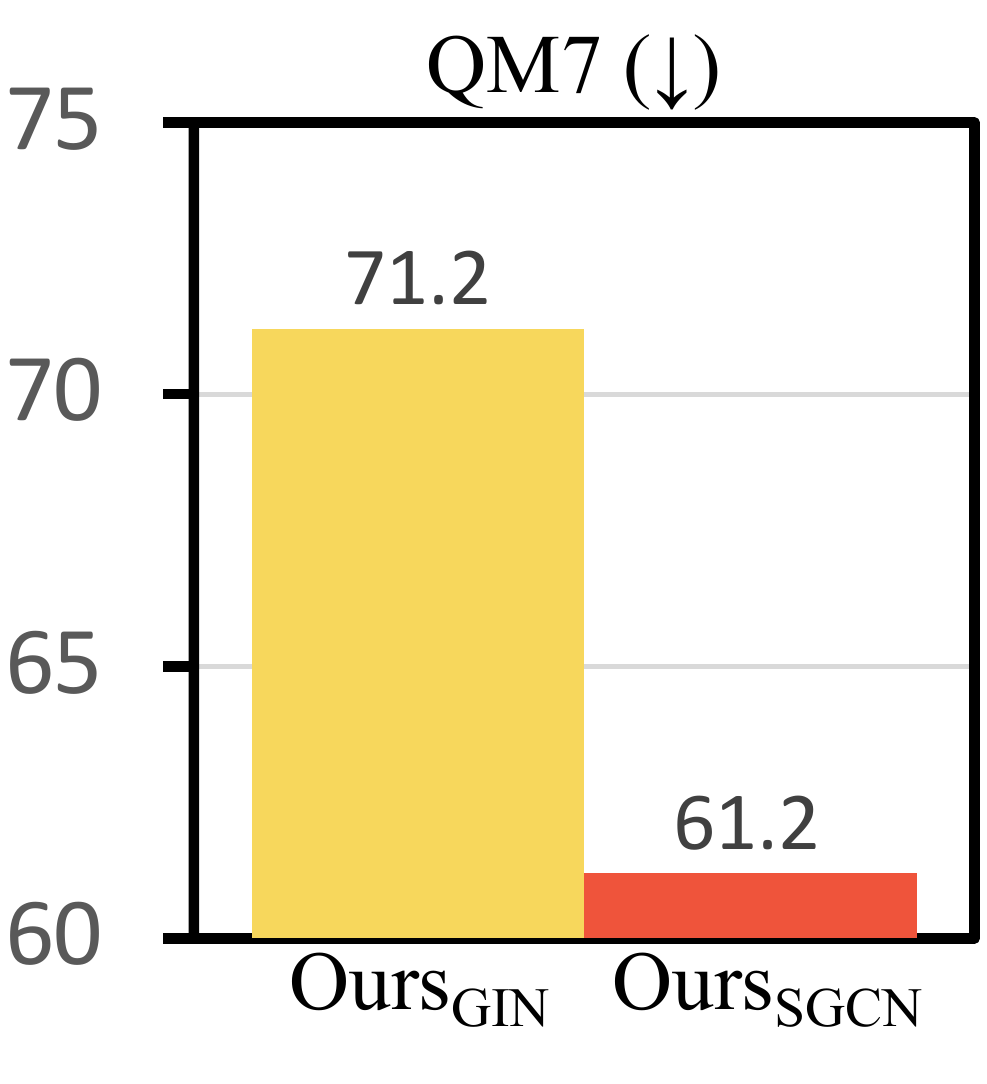}
    \vspace{-5mm}
    \caption{Comparison of our model where the GNN branch is replaced with 3D GNN.}
    
\label{fig:3d_gnn}
  \end{minipage}
\end{wrapfigure}

\paragraph{Change GNN branch to 3D-aware GNN}
To observe the effect of the choice of GNN architecture for GNN branch, we conduct additional experiments with 3D-aware GNN. We changed our GNN branch from GIN to SGCN~\citep{danel2020spatial} that utilizes 3d coordinates of input molecules. Please note that the two models are not pre-trained and instead trained on the downstream dataset from scratch. 
\autoref{fig:3d_gnn} shows that the model with 3D-aware GNN shows better performance than our original model on QM7 dataset whose task is closely related to 3D information. The result shows the flexibility of our framework that can be further tuned by using task-appropriate GNN architecture.

\paragraph{Other substructures as input tokens for Transformer}
Despite our model utilizes MACCS keys for the input of Transformer, our model can flexibly receive any substructure vocabulary. We conduct experiments using other substructures, that are ECFP fingerprints and Tree decomposition. \autoref{tab:structural_key} shows the results. It shows using MACCS keys achieves better performance than ECFP fingerprint in most of the downstream tasks. We speculate that MACCS keys include predefined functional groups such as the Hydroxyl group (-OH) whereas ECFP encodes local substructure around an atom (i.e., a certain radius neighborhood of an atom). In this aspect, ECFP fingerprint is similar to how GNN encodes node information and as we already utilize GNN to encode local information, ECFP would not bring new information about molecules.

\begin{table*}[ht]

\begin{center}
\caption{Comparison between different substructures applied to our model.}
\label{tab:structural_key}
\resizebox{\textwidth}{!}{
\begin{tabular}{lccccccccc}
\toprule
Method & BBBP~$\uparrow$  & BACE~$\uparrow$ & Tox21~$\uparrow$ & ToxCast~$\uparrow$  & FreeSolv~$\downarrow$ & ESOL~$\downarrow$ & Lipo~$\downarrow$ &  QM8~$\downarrow$ 
\\ 
\midrule
MACCS key & 0.934  & 0.868 & 0.818 & 0.725 & 2.00 & 0.878 & 0.582 &0.0140
\\
ECFP 4 & 0.925  & 0.869	 & 0.818 & 0.716 & 2.52 & 0.900 & 0.596 &  0.0152
\\
ECFP 6 & 0.903	 	 & 0.861 & 0.818 & 0.709  & 2.30 & 0.949 & 0.592 &  0.0151	
\\
Tree Decomposition & 0.925 &  0.848 & 0.796 & 0.715 & 2.37 & 0.885 & 0.614 &  -
\\
\bottomrule
\end{tabular}
}
\end{center}
\end{table*}

\section{Performance on MPG split}
\label{appendix:mpg}
As MPG uses different splits to GROVER on the paper, we reproduce the result of our model on the same split that MPG used for fair comparison. As shown in \autoref{tab:mpg_split}, our model outperforms MPG on most of the downstream datasets. 
\begin{table*}[ht]

\begin{center}
\caption{Comparison between MPG and our model on the split from MPG paper.}
\label{tab:mpg_split}
\resizebox{\textwidth}{!}{
\begin{tabular}{lccccccccc}
\toprule
Method & BBBP~$\uparrow$ & SIDER~$\uparrow$ & Clintox~$\uparrow$ & BACE~$\uparrow$ & Tox21~$\uparrow$ & ToxCast~$\uparrow$  & FreeSolv~$\downarrow$ & ESOL~$\downarrow$ & Lipo~$\downarrow$ 
\\ 
\midrule
MPG & 0.922  & 0.661 & \cellcolor{yellow}0.963 & 0.920 & 0.837 & \cellcolor{yellow}0.748 & \cellcolor{yellow}1.269 & 0.741 & 0.556 
\\
Ours & \cellcolor{yellow}0.943 & \cellcolor{yellow}0.665 & 0.958 & \cellcolor{yellow}0.931 & \cellcolor{yellow}0.847 & \cellcolor{yellow}0.748 & 1.278 & \cellcolor{yellow}0.719 & \cellcolor{yellow}0.546 
\\

\bottomrule
\end{tabular}
}
\end{center}
\end{table*}
\section{Others}

\paragraph{URLs for pretrained model for baselines.}
We use the following sources for pretrained models to reproduce the results.

Graphormer: \url{https://github.com/microsoft/Graphormer}

MAT: \url{https://github.com/ardigen/MAT}

GraphLoG: \url{https://github.com/DeepGraphLearning/GraphLoG}

GEM: \url{https://github.com/PaddlePaddle/PaddleHelix/tree/dev/apps/pretrained_compound/ChemRL/GEM}

\end{document}